\documentclass[journal]{IEEEtran}
\usepackage{amsmath,amsfonts}
\usepackage{algorithmic}
\usepackage{algorithm}
\usepackage{array}
\usepackage{bm}
\usepackage{textcomp}
\usepackage{stfloats}
\usepackage{url}
\usepackage{verbatim}
\usepackage{graphicx}
\usepackage{cite}
\usepackage{multirow}
\usepackage{multicol}
\usepackage{makecell}
\usepackage{booktabs}
\usepackage{threeparttable}
\usepackage[colorlinks,
            linkcolor=red,
            anchorcolor=blue,
            citecolor=green
            ]{hyperref}
\begin{document}

\title{A Survey of Embodied Learning for Object-Centric Robotic Manipulation}

\author{Ying Zheng$^\dagger$, Lei Yao$^\dagger$, Yuejiao Su, Yi Zhang, Yi Wang,~\IEEEmembership{Member,~IEEE}, Sicheng Zhao,~\IEEEmembership{Senior Member,~IEEE}, Yiyi Zhang, and Lap-Pui Chau$^\ast$,~\IEEEmembership{Fellow,~IEEE}
\thanks{The research work was partly conducted in the JC STEM Lab of Machine Learning and Computer Vision funded by The Hong Kong Jockey Club Charities Trust. This work was supported in part by the National Natural Science Foundation of China (Nos. 62106236).}
\thanks{Ying Zheng, Lei Yao, Yuejiao Su, Yi Zhang, Yi Wang and Lap-Pui Chau are with the Department of Electrical and Electronic Engineering, The Hong Kong Polytechnic University, Hong Kong, China. E-mail: \{ying1.zheng, yi-eie.wang, lap-pui.chau\}@polyu.edu.hk, \{rayyoh.yao, yuejiao.su, yi-eee.zhang\}@connect.polyu.hk.}
\thanks{Sicheng Zhao is with the BNRist, Tsinghua University, Beijing, China. E-mail: schzhao@tsinghua.edu.cn.}
\thanks{Yiyi Zhang is with the Department of Computer Science and Engineering, The Chinese University of Hong Kong, Hong Kong, China. E-mail: yiyizhang@link.cuhk.edu.hk.}
\thanks{$\dagger$ denotes equal contribution; $\ast$ denotes corresponding author.}}

\markboth{Journal of \LaTeX\ Class Files}%
{Shell \MakeLowercase{\textit{et al.}}: A Sample Article Using IEEEtran.cls for IEEE Journals}


\maketitle
\begin{abstract}
Embodied learning for object-centric robotic manipulation is a rapidly developing and challenging area in embodied AI. It is crucial for advancing next-generation intelligent robots and has garnered significant interest recently. Unlike data-driven machine learning methods, embodied learning focuses on robot learning through physical interaction with the environment and perceptual feedback, making it especially suitable for robotic manipulation. In this paper, we provide a comprehensive survey of the latest advancements in this field and categorize the existing work into three main branches: 1) Embodied perceptual learning, which aims to predict object pose and affordance through various data representations; 2) Embodied policy learning, which focuses on generating optimal robotic decisions using methods such as reinforcement learning and imitation learning; 3) Embodied task-oriented learning, designed to optimize the robot's performance based on the characteristics of different tasks in object grasping and manipulation. In addition, we offer an overview and discussion of public datasets, evaluation metrics, representative applications, current challenges, and potential future research directions. A project associated with this survey has been established at \url{https://github.com/RayYoh/OCRM_survey}.
\end{abstract}


\section{Introduction}
\label{sec:introduction}
\IEEEPARstart{D}{uring} the previous decade, remarkable progress has been made in machine learning research centered on deep learning, revolutionizing various fields such as computer vision \cite{he2016deep,dosovitskiy2020image} and natural language processing \cite{vaswani2017attention,devlin2019bert}. Traditional machine learning methods rely on training models using pre-constructed datasets for pattern recognition and prediction. However, these datasets are primarily derived from static sources like images, videos, and texts, which may limit their applicability and effectiveness.

Embodied learning, serving as the cornerstone of embodied AI, stands in stark contrast to traditional machine learning. It emphasizes knowledge acquisition through physical interactions and practical experiences\cite{gupta2021embodied,roy2021machine}. The data sources encompass a broad spectrum, including sensory inputs, bodily actions, and immediate environmental feedback. This learning mechanism is highly dynamic, continuously refining behaviors and manipulation strategies through real-time interactions and feedback loops. Embodied learning is essential in robotics as it equips robots with enhanced environmental adaptability, enabling them to handle changing conditions and undertake more intricate and complex tasks.

While a plethora of embodied learning methods have been proposed, this survey primarily focuses on the task of object-centric robotic manipulation. The inputs for this task are data collected from sensors, and the outputs are operational strategies and control signals for the robot to perform manipulation tasks. The objective is to enable the robot to efficiently and autonomously perform various object-centric manipulation tasks while enhancing its generality and flexibility across different environments and tasks. This task is highly challenging due to the diversity of objects and manipulation tasks, the complexity and uncertainty of the environment, and challenges such as noise, occlusion, and real-time constraints in real-world applications.

\begin{figure*}[t]
	\centering
	\centerline{\includegraphics[width=0.9\linewidth]{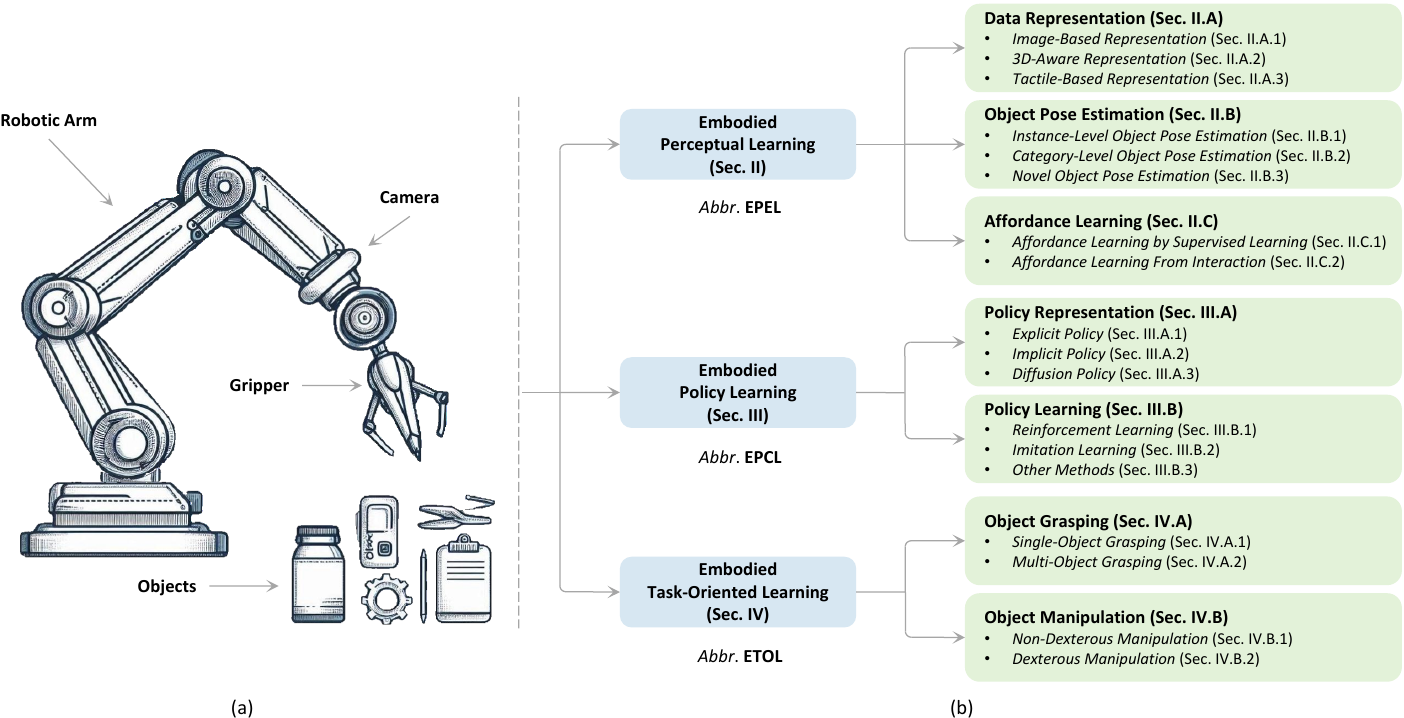}}
        \vspace{-5pt}
	\caption{An illustration of robotic manipulation system (left) and the typology of embodied learning methods for object-centric robotic manipulation (right). EPEL takes the data obtained from sensors such as cameras as input, enhancing the understanding of objects and the environment through interaction. It serves as the basis for EPCL and ETOL. EPCL utilizes the perceptual information provided by EPEL to formulate action strategies for robotic arms and end-effectors like grippers, thereby providing specific operational capabilities for ETOL. ETOL integrates EPEL and EPCL, learning to perform diverse tasks based on the characteristics of different objects. These three closely related learning processes work together to enable robots to accomplish complex tasks.}
	\label{fig:structure}
\end{figure*}

Fig. \ref{fig:structure} (a) illustrates a typical robotic manipulation system. It features a robotic arm equipped with sensors like cameras and end-effectors such as grippers, enabling it to manipulate a wide range of objects. The system's intelligence revolves around three key aspects, corresponding to the three types of embodied learning methods depicted in Fig. \ref{fig:structure} (b). 1) Advanced perception capabilities, which involve utilizing data captured by different sensors to understand the target object and external environment; 2) Precise policy generation, which entails analyzing the perceived information to make optimal decisions; 3) Task-orientation, which ensures the system can adapt to specific tasks by optimizing the execution process for maximum effectiveness.

In recent years, extensive research has been conducted around those above three key aspects, particularly with the flourishing of Large Language Models (LLMs) \cite{zhao2023survey}, Neural Radiance Fields (NeRFs) \cite{mildenhall2021nerf}, Diffusion Models \cite{ho2020denoising}, and 3D Gaussian Splatting \cite{kerbl20233d}, leading to a host of innovative solutions. However, there is a notable absence of a comprehensive survey that encapsulates the latest research in this rapidly evolving field. This motivates us to write this survey to systematically recap the cutting-edge advancements and summarize the encountered challenges, along with the prospective research directions.

\subsection{Comparison with Recent Surveys}
Over the past few years, many survey articles have emerged on embodied AI and robot learning, addressing various domains like navigation \cite{gervet2023navigating}, planning \cite{guo2023recent}, grasping \cite{du2021vision}, and manipulation \cite{han2023survey}. Table \ref{tab:surveys} summarizes some of the recent relevant surveys in this field. The survey paper by Cong et al. (2021) \cite{cong2021comprehensive} is the most closely related work to ours. Their focus is on 3D vision-based robotic manipulation and primarily reviews research on 3D visual perception up to 2021. In contrast, our work is not limited to 3D visual perception methods; we also systematically summarize and categorize representation methods based on images, 3D-aware techniques, and tactile sensing. Additionally, we provide a comprehensive introduction to critical aspects of robotic manipulation, such as policy and task-oriented learning. Notably, our survey covers a wide range of the latest research achievements published after 2021, offering a more cutting-edge and comprehensive perspective.

\begin{table}[ht]
\centering
\caption{Summary of recent surveys related to embodied AI and robot learning. RM: Robotic Manipulation; RG: Robotic Grasping; RL: Reinforcement Learning.}
\resizebox{\linewidth}{!}{
\begin{tabular}{lcl}
\toprule
{\textbf{Author}} & {\textbf{Year}} & {\textbf{Short Description}}\\
\midrule
Huang\cite{huang2016recent} & 2016 & Datasets of RM\\
Yamanobe\cite{yamanobe2017brief} & 2017 & Affordance in RM\\
Jin\cite{jin2018robot} & 2018 & Robot manipulator control using neural networks\\
Fang\cite{fang2019survey} & 2019 & Imitation learning for RM\\
Billard\cite{billard2019trends} & 2019 & Trends and challenges in RM\\
Kleeberger\cite{kleeberger2020survey} & 2020 & Machine learning for vision-based RG\\
Du\cite{du2021vision} & 2020 & Vision-based RG\\
Kroemer\cite{kroemer2021review} & 2020 & Machine learning for RM\\
\underline{Cong\cite{cong2021comprehensive}} & \underline{2021} & \underline{3D vision-based RM}\\
Zhu\cite{zhu2021deep} & 2021 & Deep learning for embodied visual navigation\\
Cui\cite{cui2021toward} & 2021 & Adaptability of learned RM\\
Zhu\cite{zhu2022challenges} & 2021 & RM of deformable objects\\
Mohammed\cite{mohammed2022review} & 2022 & RL-based RM in cluttered environments\\
Suomalainen\cite{suomalainen2022survey} & 2022 & RM in contact\\
Duan\cite{duan2022survey} & 2022 & Simulators for embodied AI\\
Francis\cite{francis2022core} & 2022 & Embodied vision-language planning\\
Zhang\cite{zhang2022robotic} & 2022 & Traditional and recent methods for RG\\
Xie\cite{xie2023learning} & 2022 & Learning-based RG\\
Gervet\cite{gervet2023navigating} & 2022 & Real-world empirical study for robot navigation\\
Han\cite{han2023survey} & 2023 & RL for RM\\
Tian\cite{tian2023data} & 2023 & RG for unknown objects\\
Newbury\cite{newbury2023deep} & 2023 & Deep learning approaches to grasp synthesis\\
Guo\cite{guo2023recent} & 2023 & Task and motion planning for robotics\\
Zare\cite{zare2023survey} & 2023 & Imitation learning\\
Xiao\cite{xiao2023robot} & 2023 & Foundation models for robot learning\\
Weinberg\cite{weinberg2024survey} & 2024 & Learning approaches for in-hand RM\\
Chen\cite{chen2024deep} & 2024 & Generative models for offline policy learning\\
Ma\cite{ma2024survey} & 2024 & Vision-language-action models for
embodied AI\\
Xu\cite{xu2024survey} & 2024 & Foundation models for robot planning and control\\
\midrule
\textbf{Ours} & 2024 & Embodied learning methods for object-centric RM\\
\bottomrule
\end{tabular}
}
\label{tab:surveys}
\end{table}

\subsection{Text Organization}
This paper presents a comprehensive survey of embodied learning methods for object-centric robotic manipulation, encompassing three main domains and seven sub-directions. The three domains are embodied perceptual learning (Sec. \ref{sec:embodiedperceptuallearning}), embodied policy learning (Sec. \ref{sec:embodiedpolicylearning}), and embodied task-oriented learning (Sec. \ref{sec:embodiedtaskorientedlearning}). The seven sub-directions include data representation (Sec. \ref{sec:datarepresentation}), object pose estimation (Sec. \ref{sec:objectposeestimation}), affordance learning (Sec. \ref{sec:affordancelearning}), policy representation (Sec. \ref{sec:policyrepresentation}), policy learning (Sec. \ref{sec:policylearning}), object grasping (Sec. \ref{sec:objectgrasping}), and object manipulation (Sec. \ref{sec:objectmanipulation}). We also extensively cover the commonly used datasets and evaluation metrics (Sec. \ref{sec:datasetsandevaluationmetrics}), along with several representative applications within this field (Sec. \ref{sec:applications}). Additionally, we delve into the primary challenges and provide insights into potential future research directions (Sec. \ref{sec:challengesandfuturedirections}).

\section{Embodied Perceptual Learning}
\label{sec:embodiedperceptuallearning}
To perform object-centric robotic manipulation, the robot must first learn to perceive the target object and its surrounding environment, which involves data representation, object pose estimation, and affordance learning. In this section, we will provide a comprehensive overview of these works.

\subsection{Data Representation}
\label{sec:datarepresentation}
In object-centric robotic manipulation, robots utilize various sensors to perceive their surroundings. These encompass visual sensors like RGB and depth cameras, which capture color images and depth maps; LiDARs, which create high-resolution 3D point clouds through distance measurements; and tactile sensors, which detect forces during grasping and pressure distribution on contact surfaces. The data collected by these sensors come in different forms, leading to various representations tailored to specific solutions. Next, we will introduce three primary types of data representation approaches: image-based representation, 3D-aware representation, and tactile-based representation.

\subsubsection{Image-Based Representation}
\begin{figure}[t]
	\centering
	\centerline{\includegraphics[width=1.0\linewidth]{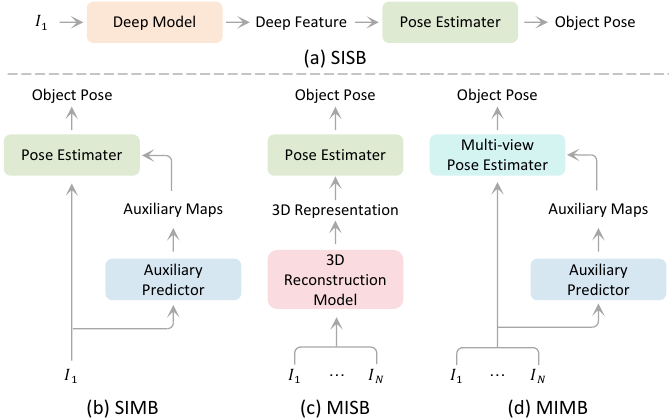}}
	\caption{Conceptual comparison of four image-based representation frameworks. SISB: Single-Image Single-Branch; SIMB: Single-Image Multi-Branch; MISB: Multi-Image Single-Branch; MIMB: Multi-Image Multi-Branch.}
	\label{fig:imagebasedrepresentation}
\end{figure}

This line of work primarily focuses on constructing effective representations solely from RGB images, thereby providing a robust foundation for subsequent tasks in robotic manipulation, such as object pose estimation. Depending on the number of input images and variations in network architecture, existing methods can be broadly categorized into four types: single-image single-branch (SISB) \cite{tekin2018real}, single-image multi-branch (SIMB) \cite{zhai2023monograspnet}, multi-image single-branch (MISB) \cite{liu2023rgbgrasp}, and multi-image multi-branch (MIMB) \cite{an2023rgbmanip}, as illustrated in Fig. \ref{fig:imagebasedrepresentation}.

(a) As depicted in Fig. \ref{fig:imagebasedrepresentation} (a), the SISB methods take a single RGB image as input, with a streamlined network architecture featuring a single main pathway. It conventionally employs deep learning models like CNNs to extract deep features from the source image, which are then fed into a pose estimator to generate the essential object pose information for robotic manipulation. SISB incorporates a typical approach to deep feature representation within an end-to-end network framework. Despite its speed and simplicity, the SISB's limitation in expressing objects' 3D geometric information may result in subsequently coarser object pose estimation.

(b) To overcome the limitations of SISB, SIMB methods introduce extra network branches alongside the main pathway, as shown in Fig. \ref{fig:imagebasedrepresentation} (b). These additional branches are designed to capture richer auxiliary information. For instance, MonoGraspNet \cite{zhai2023monograspnet} combines a keypoint network and a normal network to produce keypoint heatmaps and normal maps, respectively. It provides a more robust intermediate representation, improving pose estimation accuracy. However, this method relies heavily on the prediction accuracy of the additional branches. Due to the inherent limitations of making predictions based on a single image, errors are inevitably introduced in the generated intermediate representations. These errors can amplify the adverse effect on subsequent processing steps and increase uncertainty in robotic manipulation tasks.

(c) Owing to the lack of scale information in a single image, accurately estimating the 3D geometric information of objects is quite challenging. Therefore, a lot of research has focused on exploring methods that use multiple images to address this constraint. Among these approaches, the MISB framework has received significant attention. As shown in Fig. \ref{fig:imagebasedrepresentation} (c), this framework aims to use multiple images for 3D reconstruction to recover depth information of the scene \cite{kerr2023evo,agrawal2024clear}, which in turn facilitates the generation of efficient 3D representations. Specifically, the depth recovery can be achieved through advanced techniques such as NeRFs \cite{mildenhall2021nerf} or Gaussian Splatting \cite{kerbl20233d}.

(d) Unlike MISB, MIMB aims to directly generate multi-view image representations from images captured by a robot at multiple positions, bypassing the phase of 3D reconstruction. As illustrated in Fig. \ref{fig:imagebasedrepresentation} (d), the MIMB methods incorporate additional predictors to acquire extra information, compensating for the lack of 3D information and enhancing the robot's scene perception. For example, RGBManip \cite{an2023rgbmanip} introduces a multi-view active learning method and utilizes the segmentation maps produced by the SAM model \cite{kirillov2023segment} to provide enhanced representations for the multi-view pose estimator.

\subsubsection{3D-Aware Representation}
\begin{figure}[htbp]
	\centering
	\centerline{\includegraphics[width=0.9\linewidth]{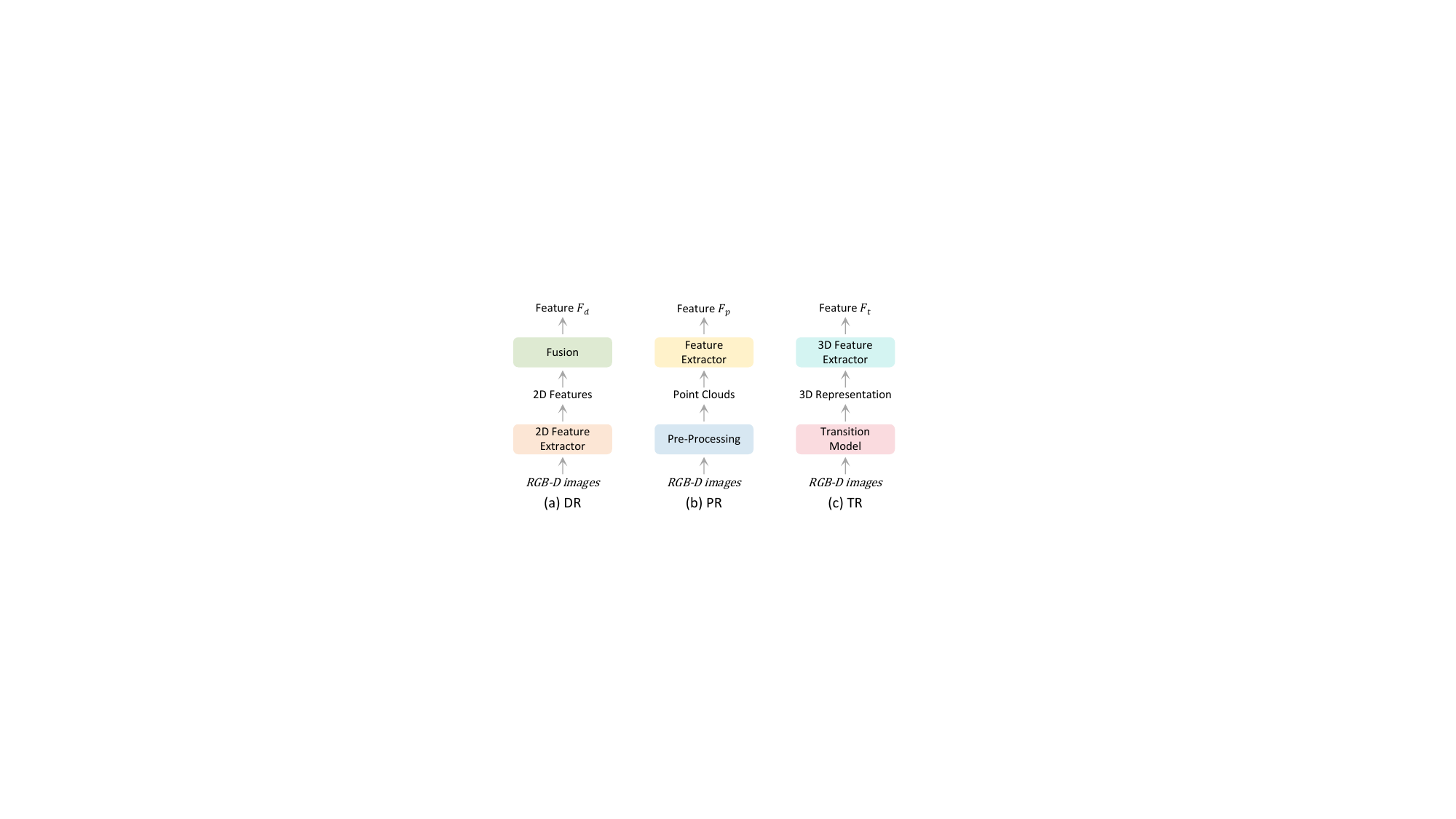}}
	\caption{Conceptual comparison of three 3D-aware representation frameworks. DR: Depth-based Representation; PR: Point cloud-based Representation; TR: Transition-based Representation.}
	\label{fig:3dawarerepresentation}
\end{figure}

This section explores 3D-aware representation, which usually takes RGB-D images as input. Existing methods fall into three categories based on the representations they generate: depth-based representation (DR), point cloud-based representation (PR), and transition-based representation (TR), as shown in Fig. \ref{fig:3dawarerepresentation}.

(a) The DR methods usually employ a network to extract 2D features from RGB-D images simultaneously, as illustrated in Fig. \ref{fig:3dawarerepresentation} (a). Some use these extracted features directly for subsequent tasks \cite{redmon2015real,lenz2015deep}, which typically necessitate posterior refinement. For example, Lenz \textit{et al.} \cite{lenz2015deep} introduced a two-stage cascade network architecture, where the first network efficiently filters out numerous unlikely grasps generated upon extracted features, and the second network concentrates on evaluating the detections from the first network. Another line of studies \cite{schwarz2018rgb,kumra2017robotic} utilizes a two-stream network to independently extract 2D features from RGB and depth images. Subsequently, these features are combined or fused to generate the final feature $F_d$ for downstream tasks.

(b) Instead of directly extracting features on RGB-D images, PR methods first create point clouds through pre-processing, as depicted in Fig. \ref{fig:3dawarerepresentation} (b). Previous approaches for processing point clouds converted from RGB-D images \cite{varley2017shape} often involve voxelizing the point clouds and utilizing 3D convolutional neural networks to extract features. However, such approaches are inefficient in terms of memory usage. The introduction of PointNet \cite{qi2017pointnet}, a network architecture designed explicitly for point clouds, has revolutionized the field. Many methods \cite{liang2019pointnetgpd,zhong20233d} now prefer to leverage PointNet-like frameworks that enable direct feature extraction from individual points in the point cloud, followed by task-specific modules customized for different objectives.

(c) Fig. \ref{fig:3dawarerepresentation} (c) presents the framework of TR works \cite{yang2018dense,goyal2023rvt} that focus on improving the model's understanding of 3D geometry by translating the input RGB-D data into 3D representations such as occupancy fields, NeRFs, or 3D Gaussians. For example, Ref. \cite{ze2023gnfactor} involves converting RGB-D data into a voxel representation, using a voxel encoder to create a 3D feature volume. This volume is then employed to construct a neural radiance field to model the 3D space and predict robot actions. Refs. \cite{lu2024manigaussian,li2024object} projects RGB-D data into dense point clouds or voxelized point clouds, which are the foundation for placing 3D Gaussians within the scene and enhancing support for robotic manipulation tasks.

\subsubsection{Tactile-Based Representation}
Tactile sensing acquires crucial force and positional information, allowing the robot to perceive contact with objects and subtle surface changes sensitively. This information is vital for enhancing the robot's capacity to perform complicated tasks and improving its operational accuracy and adaptability.

The field of tactile sensing technologies is diverse, with examples such as Gelsight \cite{yuan2017gelsight}, DIGIT \cite{lambeta2020digit}, and AllSight \cite{azulay2023allsight}. These sensors can capture various tactile information such as contact positions, normal forces, tangential forces, and torques. The representation methods for this data also vary. One common representation is time sequences obtained through multiple samplings of tactile feedback within a specific time window \cite{pham2017hand,lee2020making}. These sequences can be converted into feature vectors using neural networks like LSTM \cite{hochreiter1997long}, which simplifies the processing in subsequent models. Another form of representation is the tactile image \cite{calandra2017feeling,guzey2023dexterity}, which presents tactile information visually in an intuitive format similar to a standard RGB image and can be directly processed using CNN for feature extraction. Additionally, tactile data can be integrated with other modalities, such as vision and audio, to create a multimodal representation \cite{calandra2018more,gao2022objectfolder}, providing a comprehensive understanding of the environment and objects.

Furthermore, creating high-quality tactile representations often requires extensive training data. However, gathering tactile data is more time-consuming than visual data. To overcome this challenge, researchers have proposed leveraging technologies like NeRF or GANs to generate tactile data \cite{wang2022tacto,xu2023efficient} or building simulation environments to imitate tactile experiences \cite{zhong2023touching,dou2024tactile}. With the continuous development of these techniques, we anticipate that tactile-based representations will play an even more significant role in robotic manipulation.

\subsubsection{Discussion} Image-based representation minimizes sensor requirements but is limited by relying solely on RGB image information. 3D-aware representation leverages both image and depth data to provide a more robust representation for learning tasks. Tactile-based representation serves as a supplementary method, further enhancing the robot's perception abilities. Future research should focus on combining these methods to fully exploit their respective strengths.

\subsection{Object Pose Estimation}
\label{sec:objectposeestimation}
Grasp detection, an essential component of robotic manipulation, relies on accurate object pose estimation as a crucial step \cite{liu2024deep}. The precision of pose estimation significantly affects the robot's ability to successfully grasp target objects, emphasizing the need to develop robust and efficient pose estimation algorithms. Based on the type of predicted output, there are two main categories of object pose estimation methods: 2D planar pose estimation \cite{de2021effective} and 6D pose estimation in 3D space \cite{tejani2017latent,shugurov2021dpodv2}. The former predicts the object's position in the 2D plane and a 1D rotation angle, primarily employed for manipulating objects within a 2D plane. An example application for this method is product sorting in industrial assembly lines, where robotic arm grippers are typically positioned above the sorting platform and utilize a vertical downward angle to grasp target objects. The latter predicts the object's 6DoF (Degrees of Freedom), including 3D rotation and 3D translation, which can fully describe the object's position and orientation in 3D space. Compared to 2D planar pose estimation, 6D pose estimation has a broader range of applications, allowing robotic arms to manipulate objects from any angle.

Most existing work focuses on the 6D object pose estimation, which can be divided into three categories: instance-level, category-level, and novel object pose estimation.

\subsubsection{Instance-Level Object Pose Estimation (ILOPE)}
It refers to estimating the pose of a specific instance of an object, such as a particular cup. Existing methods typically require detailed prior knowledge of the object's shape and appearance, which a textured CAD model can furnish. Since these methods conduct training on specific samples of target objects, the trained models are object-specific.

The ILOPE problem can be formulated as Eq.\ref{eq:ilope}: Given a set of $\textit{N}_o$ objects $\mathbf{O}=\{\mathbf{o}_i|i=1,2,\dots,\textit{N}_o\}$, along with their corresponding 3D models $\mathbf{M}=\{\mathbf{m}_i|i=1,2,\dots,\textit{N}_o\}$, the objective is to learn a model $\mathbf{\Phi}$ to estimate the transformation matrix $\mathbf{T}$ for each object instance \textit{S} that is present in a given RGB or RGB-D image \textit{I}. This transformation $\mathbf{T}$ consists of a 3D rotation $\mathbf{R} \in SO(3)$ and a translation component $\textbf{t} \in \mathbb{R}^3$, which can map the target \textit{S} to the camera coordinate system.
\begin{equation}
\begin{split}
\mathbf{T} \gets \mathbf{\Phi} (\textit{I}~|~\mathbf{O},\mathbf{M}).
\end{split}
\label{eq:ilope}
\end{equation}

Significant research has been conducted to estimate the pose of objects at the instance level. Some methods utilize deep neural networks to directly regress the 6D pose of objects, such as PoseCNN \cite{xiang2017posecnn} and CDPN \cite{li2019cdpn}. However, these methods may still require post-processing optimization \cite{wang2019densefusion,iwase2021repose} to achieve better prediction results, as they are relatively simple. Another class of methods involves learning 2D-3D or 3D-3D correspondences using keypoints \cite{xu20246d} and then employing a RANSAC-based PnP (Perspective-n-Point) algorithm \cite{gao2003complete,lepetit2009ep} to generate pose estimation results. Furthermore, template matching \cite{dang2024match} or feature point voting \cite{zhou2023deep} are promising approaches for 6D object pose estimation.

The above methods have the advantage of yielding highly accurate pose estimation results. However, they require training for each instance, which makes them unsuitable for handling large-scale and diverse sets of objects.

\subsubsection{Category-Level Object Pose Estimation (CLOPE)}
It involves estimating the pose of objects belonging to predefined categories, such as cups. Existing methods for this task generally do not rely on training on specific instances of objects. Instead, they perform pose estimation using certain features within or across object classes. These methods do not require a 3D model for each instance, which is particularly beneficial when the exact shape and appearance of the objects are not known in advance.

Formally, the CLOPE problem can be stated as Eq.\ref{eq:clope}: Given a set of $\textit{N}_c$ object categories $\mathbf{C}=\{\mathbf{c}_i|i=1,2,\dots,\textit{N}_c\}$ and a set of objects $\mathbf{O}$ belonging to different categories, the goal is to learn a model $\mathbf{\Phi}$ to estimate the transformation matrix $\mathbf{T}$ for each object instance \textit{S} that appears in the observed RGB or RGB-D image \textit{I} and belongs to category $\mathbf{c}_k$. In this case, the 3D model of each object is not available.
\begin{equation}
\begin{split}
\mathbf{T} \gets \mathbf{\Phi} (\textit{I}~|~\mathbf{O},\mathbf{C}).
\end{split}
\label{eq:clope}
\end{equation}

To estimate object pose at the category level, Wang \textit{et al.} \cite{wang2019normalized} introduced NOCS (Normalized Object Coordinate Space), a coordinate system based on the object category. NOCS encodes the pose and size of the object as a normalized coordinate vector, and then the correspondence between observed pixels and NOCS can be directly inferred with a neural network. Chen \textit{et al.} \cite{chen2021sgpa} utilized the structured prior of the object category to guide pose adaptation and employed a transformer-based network to model the global structural similarity between the object instance and the prior.
These methods are mainly suitable for the pose estimation of rigid objects \cite{hai2023rigidity,corsetti2023revisiting}. However, they are not effectively generalized for articulated objects due to the complexity of articulated object poses, which involve not only translation and rotation but also various joint movements. To address category-level articulation pose estimation (CAPE), Li \textit{et al.} \cite{li2020category} expanded upon NOCS and introduced ANCSH (Articulation-aware Normalized Coordinate Space Hierarchy), a category-level representation method tailored for articulated objects. Additionally, Liu \textit{et al.} \cite{liu2022toward} proposed a real-world task setting called CAPER (CAPE-Real), which can handle multiple instances and diverse kinematic structures.

The aforementioned methods all estimate an object's pose under the assumption that the object category is known. They typically train models using datasets of known object categories and then perform pose estimation on new instances of the object. These methods enable generalization within the predefined object categories, but they are not capable of handling new object categories.

\subsubsection{Novel Object Pose Estimation (NOPE)}
It has emerged as a highly active research area in recent years to estimate the pose of novel objects from previously unseen categories during training. In this case, instance-level 3D models and category-level prior information are unavailable, but we can take reference images of the target object as an aid. This problem can be formalized as Eq.\ref{eq:nope}: Given one or multiple test images \textit{I} along with several reference images $\textit{I}_r$ associated with the target object, the objective is to learn a model $\mathbf{\Phi}$ to estimate the transformation matrix $\mathbf{T}$ within the test images by leveraging the visual information from the reference images.
\begin{equation}
\begin{split}
\mathbf{T} \gets \mathbf{\Phi} (\textit{I}~|~\textit{I}_r).
\end{split}
\label{eq:nope}
\end{equation}

In this field, classic methods usually employ image matching \cite{liu2022gen6d,nguyen2024gigapose} or feature matching \cite{sun2022onepose,goodwin2022zero} techniques and subsequently perform pose estimation on new object instances. For example, Liu \textit{et al.} \cite{liu2022gen6d} developed Gen6D, a novel 6D pose estimation method that integrates an object detector, viewpoint selector and pose refiner, enabling the inference of the 6D pose of unseen objects without relying on 3D models. Goodwin \textit{et al.} \cite{goodwin2022zero} proposed a method based on a self-supervised vision transformer and semantic correspondence to achieve zero-shot object pose estimation.

Recently, the research community has been increasingly focused on utilizing large models to enhance the generalization capability of deep models for the NOPE task. Lin \textit{et al.} \cite{lin2024sam} introduced the SAM-6D approach, which employs the powerful semantic segmentation capabilities of Segment Anything Model (SAM) \cite{kirillov2023segment} to generate potential object proposals. Simultaneously, Wen \textit{et al.} \cite{wen2024foundationpose} investigated methods to integrate LLMs with contrastive learning, significantly improving model generalization by training on large-scale synthetic datasets. The primary advantage of these methods is that they can handle new object categories, thereby enhancing their generalizability and applicability in a broader range of real-world scenarios. However, it should be noted that large models usually require more training data and computational resources, which could be a potential limitation.

\subsubsection{Discussion} These three types of pose estimation methods each have specific application scenarios and advantages and disadvantages: ILOPE offers high accuracy but is only suitable for known objects; CLOPE has a wide range of applicability but relatively lower accuracy; NOPE is highly flexible but faces significant challenges in accuracy and robustness.

\subsection{Affordance Learning}
\label{sec:affordancelearning}

\begin{figure}[t]
	\centering
	\centerline{\includegraphics[width=1.0\linewidth]{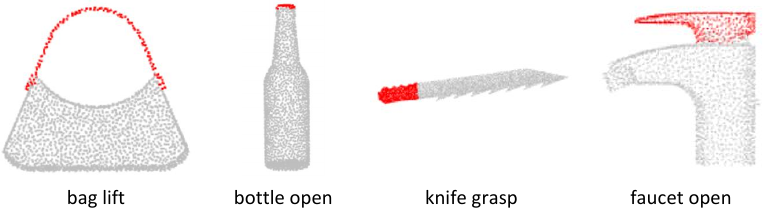}}
	\caption{Visualization of four representative affordance prediction examples from the dataset provided by \cite{li2024laso}, including bag lift, bottle open, knife grasp, and faucet open. The affordance ground truth labels are highlighted in \textcolor{red}{red}.}
	\label{fig:affordance}
\end{figure}

Once the estimated object pose is obtained, the next step involves identifying potential interactive regions of the object as shown in Fig. \ref{fig:affordance}, a process known as affordance learning \cite{gibson1977theory}. As a crucial component of robotic manipulation, affordance learning enables robots to comprehend the object's functionality and potential actions. Based on the data source, affordance learning can be categorized into two types: affordance learning by supervised learning and affordance learning from interaction.

\subsubsection{Affordance Learning by Supervised Learning}
In order to make robots understand object manipulation, various methods have been proposed that utilize static data to learn affordances \cite{koppula2015anticipating,do2018affordancenet}. For example, AffordanceNet \cite{do2018affordancenet} considered human-annotated RGB images from public datasets as input and simultaneously performed object localization and affordance prediction through two distinct branches. Specifically, this method assigned a probable affordance label to each pixel within the predicted object, effectively making it a part of a semantic segmentation task. Additionally, Nagarajan \textit{et al.} \cite{nagarajan2019grounded} utilized interaction hotspot maps to depict object affordances and trained their model on large-scale human-object interaction video datasets.

While the aforementioned methods have shown promising results on static datasets, they have not explored applying learned affordances to robotic manipulation tasks. To bridge this gap, VRB \cite{bahl2023affordances} incorporated a trajectory prediction model to extract affordances from egocentric videos and integrated the resulting model into various robot learning frameworks. To improve generalization to unseen objects, Robo-ABC \cite{ju2024robo} emphasized semantic correspondence and has successfully implemented its model on real-world platforms for grasping novel objects. Additionally, RAM \cite{kuang2024ram} developed a retrieval-based architecture that lifts 2D affordances to 3D, enabling embodiment-agnostic robotic manipulation. This framework introduced a hierarchical retrieval pipeline to transfer actionable knowledge from out-of-domain data to specific target domains. To overcome the constraints of closed-set affordance learning, OpenAD \cite{nguyen2023open} measured the similarity between language-based affordance labels and point-wise high-dimensional features and extended affordance learning to an open-vocabulary context. 

\subsubsection{Affordance Learning From Interaction}
Rather than relying on supervised learning from static data, affordance learning from interaction framework seeks to gather training data through simulations. This method allows the system to learn from interactions, providing it with essential prior knowledge for real-world deployment. As a pioneer in this field, Where2Act \cite{mo2021where2act} employed self-supervised interaction for articulated 3D objects, which uses single-frame images or partial point clouds as observations in the SAPIEN \cite{xiang2020sapien} simulator. But AdaAfford \cite{wang2022adaafford} identified that this paradigm ignores hidden kinematic uncertainties that lead to inaccurate affordances. To address this, AdaAfford proposed a method that involves sampling multiple test-time interactions to facilitate rapid adaptation. Building on similar concepts, DualAfford \cite{zhao2023dualafford} expanded the interactive learning framework to dual-gripper manipulation to broaden the robot's manipulation capabilities. Nevertheless, relying on random interactions for data collection makes these methods sample-inefficient. ActAIM \cite{wang2023self} tackled this with a clustering-based strategy and a generative model to improve interaction diversity and data quality. Additionally, IDA \cite{mazzaglia2024information} put forward an information-driven method for affordance discovery to boost interaction efficiency. Where2Explore \cite{ning2024where2explore} generalized affordance recognition to novel instances and even various object categories by leveraging local geometries for actionable parts.

It is important to note that all the aforementioned methods operate under the assumption of noiseless visual information, which is often unrealistic. In response, Ling \textit{et al.} \cite{ling2024articulated} introduced a coarse-to-fine architecture to reduce point cloud noise and improve the affordance learning performance. Beyond focusing solely on single-object affordances, Cheng \textit{et al.} \cite{wu2024learning} incorporated realistic physical constraints within environments and employed a data-efficient contrastive learning method to acquire environment-aware affordances, even under occlusions. RLAfford \cite{geng2023rlafford}, in contrast to prior work limited by predefined affordance primitives, integrated reinforcement learning to facilitate end-to-end affordance learning. Specifically, they considered contact maps of interest during the RL process as visual affordances and seamlessly adapted the architecture to various manipulation tasks.

\subsubsection{Discussion} Current supervised affordance learning methods are limited by their focus on specific domain data or tasks, while interaction-based approaches are constrained by sample inefficiency. Future research should investigate how to design effective frameworks that can harness the vast potential of internet-scale data and rapidly adapt to specific tasks.

\section{Embodied Policy Learning}
\label{sec:embodiedpolicylearning}
Embodied policy learning aims to empower robots with the sophisticated decision-making capabilities required to perform manipulation tasks efficiently. This section will delineate the process of embodied policy learning into two fundamental phases: policy representation and policy learning, elucidating how these techniques enable robots to accomplish predefined objectives. We summarize the key works in embodied policy learning in Table \ref{tab:policy}.

\begin{table*}[t]
	\centering
	\caption{Summary of embodied policy learning. RL: Reinforcement Learning; IL: Imitation Learning.}
	\resizebox{\linewidth}{!}{
	\begin{tabular}{ccc}
	\toprule
	{\textbf{Task}} & {\textbf{Type}} & {\textbf{Subfields \& references}}\\
	\midrule
	\multirow{4}{*}[-2ex]{Policy Representation} & \multirow{1}{*}{Explicit Policy} & \makecell[c]{Deterministic policy \cite{kalashnikov2018scalable}, Stochastic policy \cite{liang2024rapid}}\\
	\cmidrule(r){2-3}
	{} & \multirow{1}{*}{Implicit Policy} & \makecell[c]{EBMs \cite{boney2020regularizing}, Implicit behavioral cloning \cite{florence2022implicit}, IDAC \cite{yue2020implicit}, EBIL \cite{liu2020energy}}\\
	\cmidrule(r){2-3}
	{} & \multirow{2}{*}[-1ex]{Diffusion Policy} & \makecell[c]{Diffusion Policy \cite{chi2023diffusion}, Decision Diffuser \cite{ajay2022conditional}, Diffusion-QL \cite{wang2022diffusion}, HDP \cite{ma2024hierarchical}, UniDexFPM \cite{wu2024unidexfpm}, BESO \cite{reuss2023goal}}\\
	\cmidrule(r){3-3}
	{} & {} & \makecell[c]{\textit{Incorporating language instructions}: MDT \cite{reuss2024multimodal}, Lan-o3dp \cite{li2024language}}\\
	\midrule
	\multirow{8}{*}[-5ex]{Policy Learning} & \multirow{2}{*}[-1ex]{RL} & \makecell[c]{ViSkill \cite{huang2023value}, RMA$^2$ \cite{liang2024rapid}, SAM-RL \cite{lv2022sam}, Offline RL \cite{mandlekar2021matters, levine2020offline}, Demonstration-guided RL \cite{huang2023guided}}\\
	\cmidrule(r){3-3}
	{} & {} & \makecell[c]{\textit{Rewards function learning}: Text2reward \cite{xie2023text2reward} and EUREKA \cite{ma2023eureka}}\\
	\cmidrule(r){2-3}
	{} & \multirow{4}{*}[-2ex]{IL} & \makecell[c]{DMPs \cite{schaal2005learning}, DAgger \cite{ross2011reduction}, SpawnNet \cite{lin2023spawnnet}, ACT \cite{zhao2023act}}\\
	\cmidrule(r){3-3}
	{} & {} & \makecell[c]{\textit{Scaling up demonstration data}: MimicGen \cite{mandlekar2023mimicgen}, Bridge Data \cite{ebert2021bridge}, Open X-embodiment \cite{padalkar2023open}}\\
	\cmidrule(r){3-3}
	{} & {} & \makecell[c]{\textit{Learning from human videos}: Vid2Robot \cite{jain2024vid2robot}, Ag2Manip \cite{li2024ag2manip}, MPI \cite{zeng2024learning}}\\
	\cmidrule(r){3-3}
	{} & {} & \makecell[c]{\textit{Equivariant models}: NDFs \cite{simeonov2023se}, L-NDF \cite{chun2023local}, EDFs \cite{ryu2022equivariant}, EDGI \cite{brehmer2024edgi}, Diffusion-EDFs \cite{ryu2023diffusion}, SE(3)-DiffusionFields \cite{urain2023se}}\\
	\cmidrule(r){2-3}
	{} & \multirow{2}{*}[-1ex]{Other Methods} & \makecell[c]{\textit{Combination of RL \& IL}: UniDexGrasp \cite{xu2023unidexgrasp}, UniDexGrasp++ \cite{wan2023unidexgrasp++}}\\
	\cmidrule(r){3-3}
	{} & {} & \makecell[c]{\textit{LLM- or VLM-driven}: VILA \cite{hu2023look}, Grounding-RL \cite{szot2024grounding}, OpenVLA \cite{kim2024openvla}, 3D-VLA \cite{zhen20243dvla}}\\
	\bottomrule
	\end{tabular}
	}
	\label{tab:policy}
	\end{table*}

\subsection{Policy Representation}
\label{sec:policyrepresentation}
The role of policy is to model the robot's behavior by taking its observation 
as input and determining the corresponding action to execute. This process is mathematically represented as $\pi: \mathcal{O} \mapsto \mathcal{A}$, where $\mathcal{O}$ and $\mathcal{A}$ represent the observation space and action space, respectively. Policy representation is critical in embodied policy learning, as it significantly affects the robot's decision-making ability. Depending on the modeling options, policy representation is classified into explicit, implicit, and diffusion policies, regardless of whether the action space is discrete or continuous.

\subsubsection{Explicit Policy}
Explicit policies utilize a parameterized function to map a robot's current observation $\mathbf{v} \in \mathcal{O}$ directly to an action $\mathbf{a} \in \mathcal{A}$.  Typically, explicit policies are parameterized using feed-forward models like neural networks and can be either deterministic \cite{kalashnikov2018scalable} or stochastic \cite{liang2024rapid}. A deterministic policy directly predicts an action $\mathbf{a} = \pi_{\theta}(\mathbf{v})$ to execute, while a stochastic policy samples actions from an estimated distribution $\mathbf{a} \sim \pi_{\theta}(\cdot|\mathbf{v})$, where $\theta$ indicates 
parameters of the policy. Stochastic policies enhance an agent's exploration capabilities and provide greater robustness in complex, uncertain environments compared to deterministic policies.

In a discrete action space, policy representation can be transformed into an optimal action selection process from a finite set of actions. The categorical distribution is commonly used to calculate action probabilities, from which actions are sampled based on the estimated results. For instance, Zhang \textit{et al.} \cite{zhang2022learning} conceptualized the robot assembly manipulation policy as translation, rotation, and insertion primitives, with RL subsequently optimizing the policy. In continuous action spaces, a diagonal Gaussian distribution is often chosen to represent the action distribution, guided by regression losses such as mean squared error (MSE) or RL-based objectives. The policy outputs both the mean $\mu_{\theta}(\mathbf{v})$ and the standard deviation $\sigma_{\theta}(\mathbf{v})$, and actions are sampled from the resulting distribution as follows:
\begin{equation}
    \mathbf{a} = \mu_{\theta}(\mathbf{v}) + \sigma_{\theta}(\mathbf{v}) \odot \mathbf{\xi}.
    \label{eq:stochastic}
\end{equation}
Here, $\mathbf{\xi} \sim \mathcal{N}(0, \mathbf{I})$ represents a vector of Gaussian noise, and $\odot$ signifies the Hadamard product. It should be noted that, in practical applications, the logarithm of the standard deviation $\log \sigma_{\theta}(\mathbf{v})$ is typically used to prevent the standard deviation from taking on negative values.

\subsubsection{Implicit Policy}
Unlike explicit policy models, implicit policies attempt to assign value to each action by leveraging energy-based models (EBMs) \cite{boney2020regularizing, florence2022implicit}, which are also recognized as action-value functions \cite{yue2020implicit}. This paradigm learns the policy by optimizing a continuous function to find the action with minimal energy:
\begin{equation}
    \hat{\mathbf{a}} = \mathop{\arg\min}\limits_{\mathbf{a} \in \mathcal{A}} \mathcal{E}_{\theta}(\mathbf{v}, \mathbf{a}),
    \label{eq:implicit}
\end{equation}
where $\theta$ denotes the parameters of the energy function $\mathcal{E}_{\theta}$. Consequently, the problem of action prediction is effectively reformulated as an optimization problem.

Generally, given a series of expert demonstrations or online-collected trajectories denoted as $\{(\mathbf{v}_{t}, \mathbf{a}_{t})\}_{t=0}^{T}$, implicit policies are trained by an InfoNCE-style loss \cite{oord2018representation}. Once trained, stochastic optimization will be applied to identify the optimal action for implicit inference. EBIL \cite{liu2020energy} incorporated EBMs into the inverse RL architecture, utilizing the estimated expert energy as a surrogate reward. Florence \textit{et al.} \cite{florence2022implicit} further proposed an implicit behavioral cloning approach grounded in this framework and assessed its performance across various robotic task domains such as simulated pushing and bi-manual sweeping. 

\subsubsection{Diffusion Policy}
Drawing inspiration from Denoising Diffusion Probabilities Models (DDPMs) \cite{ho2020denoising}, which gradually denoise random inputs to generate data samples, diffusion policies model the policy as a conditional generative model \cite{wang2022diffusion}. This approach approximates the action distribution $\pi(\cdot|\mathbf{v})$ considering the observation $\mathbf{v}$ as a condition for producing the corresponding action $\mathbf{a}$:
\begin{equation}
    \mathbf{a}^{k-1} = \alpha(\mathbf{a}^{k}-\beta \epsilon_{\theta}(\mathbf{a}^{k}, \mathbf{v}, k))+\sigma \mathcal{N}(0, \mathbf{I}),
    \label{eq:diffusion}
\end{equation}
where $k=1,2,\dots,K$, representing the denosing iterations, $\alpha, \beta, \sigma$ are functions rely on the noise schedule, $\epsilon_{\theta}$ denotes the denosing network with parameters $\theta$, and $\mathcal{N}(0, \mathbf{I})$ represents the standard Gaussian noise. 

As concurrent work, Decision Diffuser \cite{ajay2022conditional} and Diffusion-QL \cite{wang2022diffusion} have pioneered the integration of diffusion policies into offline RL. These studies revealed that this approach yields highly expressive policy representation that surpasses traditional policy formats. While Decision Diffuser \cite{ajay2022conditional} suggested extending diffusion policies to handle high-dimensional visual observations, its current focus remains on state-based benchmarks. In contrast, Chi \textit{et al.} \cite{chi2023diffusion} proposed a novel diffusion policy tailored for vision-based robotic manipulation tasks. Their experimental findings highlighted the efficacy of diffusion policies in visuomotor policies and their superiority in managing behavioral multimodality in imitation learning. They also incorporated techniques such as receding horizon control and time-series diffusion transformers to adapt the policy for high-dimensional action spaces, resulting in more stable training. HDP \cite{ma2024hierarchical} integrated diffusion policies into a high-level planning agent for multi-task robotic manipulation, whereas UniDexFPM \cite{wu2024unidexfpm} applied diffusion policies to pre-grasp manipulation. By leveraging the conditional generative paradigm, diffusion policies are well-suited for multimodal policy learning. For example, MDT \cite{reuss2024multimodal} and Lan-o3dp \cite{li2024language} advanced multimodal policy learning by incorporating language instructions. Differently, BESO \cite{reuss2023goal} facilitated rapid inference in diffusion policies by decoupling the score model learning from the sampling process.

\subsubsection{Discussion} Explicit policies are straightforward to implement but struggle with complex tasks, while implicit policies face challenges in training stability and computational costs. Diffusion policies offer a promising alternative to provide a more expressive and robust policy representation, but how to accelerate the sampling process remains to be explored.

\subsection{Policy Learning}
\label{sec:policylearning}
After establishing a suitable policy representation, the next critical task is to train the policy $\pi$ to execute specific manipulation tasks effectively. Policy learning methods can be broadly categorized into several approaches, including Reinforcement Learning (RL) \cite{zhang2022adjacency,hu2024transforming}, Imitation Learning (IL) \cite{torabi2018behavioral,ho2016generative}, and other methods \cite{wan2023unidexgrasp++,hu2023look} that combine elements of both or introduce entirely different learning paradigms. The choice of policy learning method depends on factors such as the availability of demonstration data, task complexity, and computational resources. Each method has its advantages and challenges, and the field of embodied policy learning continues to evolve with new techniques and insights.

\subsubsection{Reinforcement Learning}
By modeling the policy learning procedure as a Markov Decision Process (MDP), RL aims to discover the optimal policy $\pi^{*}$ that can maximize expected cumulative discounted reward, formulated as:
\begin{equation}
    \mathcal{J}(\pi) = \mathbb{E}_{\tau \sim \pi}[\sum_{t=0}^{T} \gamma^{t} r_t(\mathbf{v}_t, \mathbf{a}_t)],
    \label{eq:rl}
\end{equation} 
where $\tau = \{(\mathbf{v}_t, \mathbf{a}_t)\}_{t=0}^{T}$ denotes a trajectory, with $\mathbf{v}_t$ and $\mathbf{a}_t$ representing the observation and action at time step $t$, respectively. The function $r_t$ corresponds to the reward provided as feedback from the environment after each action is taken. Here, $\gamma \in [0, 1]$ is a discount factor used to balance the importance of immediate and future rewards. Therefore, the objective of RL can be expressed as:
\begin{equation}
    \pi^{*} = \mathop{\arg\max}\limits_{\pi} \mathcal{J}(\pi)
    \label{eq:rl_objective}.
\end{equation}

As a pivotal element for decision-making, RL has been extensively investigated in robotic manipulation. Researchers from OpenAI \cite{andrychowicz2020learning} developed a sim-to-real training pipeline to enable a physical five-fingered robot hand to perform vision-based object reorientation. This pipeline initially trained the policy in simulation using Proximal Policy Optimization (PPO) \cite{schulman2017proximal} and then adapted it to physical hardware through domain randomization. It should be highlighted that PPO is a widely used on-policy RL algorithm in robotic manipulation, valued for its simplicity and effectiveness. For long-horizon surgical robot tasks, ViSkill \cite{huang2023value} introduced a novel mechanism named value-informed skill chaining to learn smooth subtask policies. To create generalizable manipulation policies adaptable to various object shapes, RMA$^2$ \cite{liang2024rapid} presented a two-stage training framework with an extra adapter training phase within PPO, enhancing the policy's robustness across diverse objects. Inspired by model-based RL \cite{fu2016one}, SAM-RL \cite{lv2022sam} proposed a sensing-aware architecture that renders images from different viewpoints and refines the learned world model by aligning these generated images with actual raw observations, demonstrating significant real-world performance. Mandlekar \textit{et al.} \cite{mandlekar2021matters} explored the impact of various design choices in offline RL \cite{levine2020offline} and made their dataset publicly available for further research. To overcome exploration challenges in RL, Huang \textit{et al.} \cite{huang2023guided} proposed demonstration-guided RL, which assigns high values to expert-preferred actions using non-parametric regression.

Beyond algorithmic enhancements, crafting the reward function remains a significant challenge in RL due to the need for domain-specific knowledge to accurately capture the task objectives. Recently, research has increasingly explored the capabilities of LLMs for reward learning. For instance, Text2reward \cite{xie2023text2reward} and EUREKA \cite{ma2023eureka} leveraged the understanding and generation capabilities of LLMs to convert natural language descriptions of goals into dense and interpretable reward codes, which can be iteratively refined by human feedback. This iterative process is crucial as it allows the reward function to evolve in response to new insights or changes in the task requirements. Consequently, this method streamlined the resolution of complex manipulation tasks, reducing reliance on manually crafted reward functions and potentially increasing the effectiveness of the learning process.

\subsubsection{Imitation Learning}
Instead of learning in a trial-and-error manner as RL, the objective of IL is to mimic the expert behavior. 
Typically, IL encompasses three primary methodologies: Behavioral Cloning (BC) \cite{torabi2018behavioral}, Inverse Reinforcement Learning (IRL) \cite{ziebart2008maximum}, and Generative Adversarial Imitation Learning (GAIL) \cite{ho2016generative}. BC is a straightforward yet effective approach that learns the policy by minimizing the mean squared error between the expert's action and the policy's predictions through supervised learning. IRL operates in a two-stage loop, initially deducing a reward function from the demonstrations, followed by policy optimization using RL techniques. GAIL
is a generative model-based method that relies on adversarial learning to develop a discriminator and an action generator simultaneously, distinguishing between the actions of an expert and those produced by the policy.

Earlier, differentiable nonlinear dynamic systems like Dynamic Movement Primitives (DMPs) \cite{schaal2005learning} were used to acquire skills from demonstrations at the trajectory level. The essence of DMPs lies in incorporating a forcing term, comprised of a set of weighted basis functions, into the system dynamics. These weights are determined through regression analysis of the desired trajectory. Despite using a limited number of parameters, the effectiveness of DMPs is constrained by the choice of basis functions. Instead, DAgger \cite{ross2011reduction} incrementally aggregated current policy interaction data with expert policy demonstrations to augment the training data. SpawnNet \cite{lin2023spawnnet} incorporated a pre-trained visual model to develop a generalizable policy for diverse manipulation tasks. Kim \textit{et al.} \cite{kim2021transformer} introduced a self-attention mechanism to filter out irrelevant information, while ACT \cite{zhao2023act} directly trained a generative transformer model on action sequences specifically for dual-arm manipulation on real-world collected data.

Given the high cost of collecting human demonstrations, there is a focus on scaling up the demonstration data. MimicGen \cite{mandlekar2023mimicgen} designed a system that inputs a few expert demonstrations and created an augmented dataset by integrating various scenes and segmented objects. Instead, initiatives like Bridge Data \cite{ebert2021bridge} and Open X-embodiment \cite{padalkar2023open} strove to compile extensive human demonstration datasets across diverse domains. In addition, some researchers explored the potential of in-the-wild data for IL, capitalizing on the availability of extensive egocentric human activity videos. Vid2Robot \cite{jain2024vid2robot} proposed an end-to-end policy learning framework by training a unified model on human video data. Recent efforts, such as Ag2Manip \cite{li2024ag2manip} and MPI \cite{zeng2024learning}, also adopted this approach to extract skills from human videos, demonstrating substantial performance in multi-task robotic manipulation. 

Equivariant models have garnered attention for their advantages of enhancing sample efficiency and generalization in IL. A notable example is the work by Simeonov \textit{et al.} \cite{simeonov2022neural}, which introduces Neural Descriptor Fields (NDFs). These fields leverage SE(3)-equivariance to represent manipulated objects and facilitate IL by searching matched poses within demonstration data. Building on this foundation, Local Neural Descriptor Fields (L-NDF) \cite{chun2023local} extended the concept by introducing shared local geometric features between objects. However, NDFs face inherent limitations that restrict their generalization to non-fixed targets. To address this, Equivariant Descriptor Fields (EDFs) \cite{ryu2022equivariant} reformulated NDFs within a probabilistic learning framework, enhancing its flexibility. Further advancements include the integration of diffusion models into EDFs, as seen in Diffusion-EDFs \cite{ryu2023diffusion}, EDGI \cite{brehmer2024edgi}, and SE(3)-DiffusionFields \cite{urain2023se}. These approaches aim to improve the model's ability to generalize across a broader range of scenarios.

\subsubsection{Other Methods}
In the domain of embodied policy learning, several innovative methods have emerged that combine the strengths of RL and IL. As a series of work, UniDexGrasp \cite{xu2023unidexgrasp} and UniDexGrasp++ \cite{wan2023unidexgrasp++} perpetuated the paradigm of teacher-student learning, aiming to develop a universal grasp policy that can effectively generalize across diverse objects and scenarios. Initially, these methods employed model-free RL algorithms to cultivate a teacher model that takes oracle states as input. Subsequently, the skills acquired by the teacher model are distilled to a student policy via IL, where the student policy solely has access to realistic observations, such as those obtained through vision.

Recent breakthroughs in LLMs and Vision-Language Models (VLMs) have sparked interests in their applications for policy learning in robotics, leveraging their capabilities in perception, reasoning, and decision-making. These models took current visual observations and language instructions as inputs to generate corresponding action sequences through trainable adapters, enabling robots to perform complex tasks and adapt to new situations \cite{hu2023look, szot2024grounding}. Notable examples include VILA \cite{hu2023look} and Grounding-RL \cite{szot2024grounding}, which use pre-trained LLMs in their policy learning methods. In contrast, OpenVLA \cite{kim2024openvla} employed pre-trained visual encoders to extract visual features, subsequently mapping them into a language embedding space. This method utilized a low-rank adaptation fine-tuning strategy to customize LLMs for robotic manipulation tasks. 3D-VLA \cite{zhen20243dvla} further extended this concept by incorporating 3D spatial observations and integrating a diffusion model for goal-aware state generation, resulting in a 3D generative world model.

\subsubsection{Discussion} Although RL and IL have shown remarkable progress in embodied policy learning, challenges remain in terms of sample efficiency, domain adaptation, and generalization. Future research needs to unleash the potential of LLMs and VLMs for building generalizable and versatile agents for robotic manipulation tasks.

\begin{figure}[t]
	\centering
	\centerline{\includegraphics[width=1.0\linewidth]{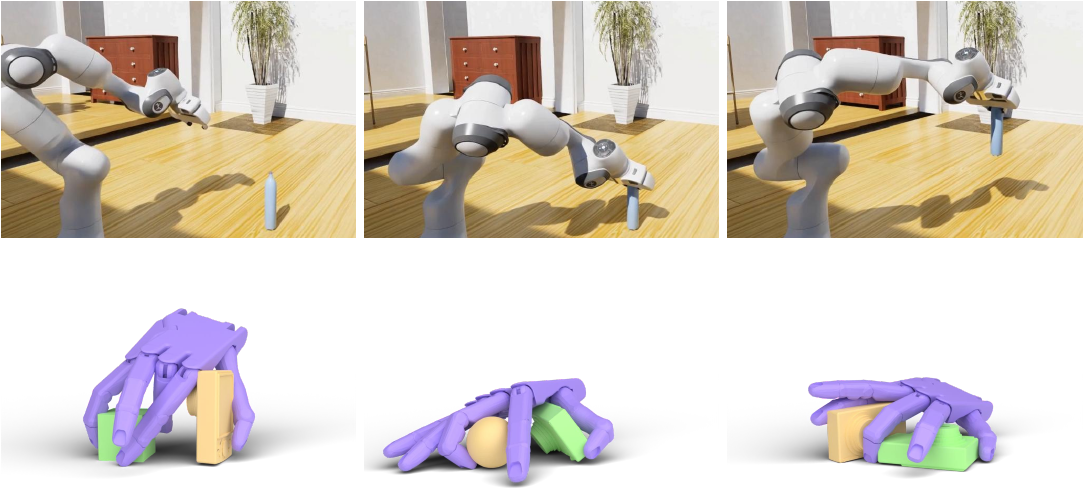}}
	\caption{Illustration of single-object grasping (top row) and multi-object grasping (bottom row). The examples are respectively from the ARNOLD benchmark \cite{gong2023arnold} and Grasp'Em dataset \cite{li2024grasp}.}
	\label{fig:objectgrasping}
\end{figure}

\section{Embodied Task-Oriented Learning}
\label{sec:embodiedtaskorientedlearning}
Embodied task-oriented learning not only involves strategic planning through powerful perception but also necessitates robots to understand how their physical attributes influence decision-making and task execution. It helps robots develop the ability to make decisions in complex and dynamic scenarios. Specifically, existing work of embodied task-oriented learning centers on two domains: object grasping and object manipulation. As shown in Table \ref{tab:embodiedtaskorientedlearning}, this section will introduce methods tailored for these two tasks, revealing how embodied learning improves the efficiency and precision of robots.

\begin{table*}[t]
\centering
\caption{Summary of embodied task-oriented learning methods. SOG: Single-Object Grasping; MOG: Multi-Object Grasping; NDM: Non-Dexterous Manipulation; DM: Dexterous Manipulation; H2R: Human-to-Robot.}
\resizebox{\linewidth}{!}{
\begin{tabular}{ccc}
\toprule
{\textbf{Task}} & {\textbf{Type}} & {\textbf{Subfields \& references}}\\
\midrule
\multirow{6}{*}[-6ex]{Object Grasping} & \multirow{4}{*}[-5ex]{SOG} & \makecell[c]{Open-loop grasping (STEM-CaRFs\cite{asif2017rgb}, FANet\cite{zhai2023fanet}, AnyGrasp\cite{fang2023anygrasp})\\closed-loop grasping (adaptive grasping\cite{marturi2019dynamic}, GG-CNN\cite{morrison2018closing}, VFAS-Grasp\cite{heppert2024ditto})}\\
\cmidrule(r){3-3}
{} & {} & \makecell[c]{\textit{Transparent object grasping}: DFNet\cite{fang2022transcg}, Dex-NeRF\cite{ichnowski2022dex}, GraspNeRF\cite{dai2023graspnerf}, NFL\cite{lee2023nfl}, TRansPose\cite{kim2023transpose}, TGF-Net\cite{yu2023tgf}}\\
\cmidrule(r){3-3}
{} & {} & \makecell[c]{\textit{Grasping in clutter}: collision-free grasping (Contact-GraspNet\cite{sundermeyer2021contact}, CaTGrasp\cite{wen2022catgrasp}, CollisionNet\cite{murali20206}, DDGC\cite{lundell2021ddgc}, GSNet\cite{wang2021graspness}, DAL\cite{wei2023discriminative})\\reposition-based grasping (push-grasping synergy\cite{xu2021efficient},  object singulation\cite{kiatos2019robust}, grasping invisible\cite{yang2020deep}, vision-language grasping\cite{xu2023joint})}\\
\cmidrule(r){3-3}
{} & {} & \makecell[c]{\textit{Dynamic object grasping}: H2R handover (wearable sensing\cite{wang2018controlling}, TLP\cite{wang2021predicting}, reactive handover\cite{yang2021reactive}, flexible handover\cite{zhang2023flexible}, GenH2R\cite{wang2024genh2r})\\human-free moving object grasping (velocity decomposition\cite{ye2018velocity}, adaptive motion generation\cite{akinola2021dynamic}, Moving GraspNet\cite{liu2023target})}\\
\cmidrule(r){2-3}
{} & \multirow{2}{*}[-1ex]{MOG} & \makecell[c]{Holistic grasping (MOG in the Plane\cite{agboh2022multi}, MOG-Net\cite{agboh2023learning}, experience forest\cite{chen2024multi}, Push-MOG\cite{aeron2023push})}\\
\cmidrule(r){3-3}
{} & {} & \makecell[c]{Independent grasping (MOG by exploiting kinematic redundancy\cite{yao2023exploiting}, MultiGrasp\cite{li2024grasp})}\\
\midrule
\multirow{3}{*}[-1.5ex]{Object Manipulation} & {NDM} & \makecell[c]{Pick-and-place \cite{berscheid2020self}, object rearrangement \cite{zhai2023sg}, kit assembly \cite{zakka2020form2fit}, \\deformable object manipulation (clothing \cite{huang2022mesh}, ropes \cite{mitrano2021learning}, and fluids \cite{li20223d}), \\articulated object manipulation (GAPartNet\cite{geng2023gapartnet}, UniDoorManip \cite{li2024unidoormanip}, PartManip\cite{geng2023partmanip})}\\
\cmidrule(r){2-3}
{} & \multirow{2}{*}[-1ex]{DM} & \makecell[c]{Trajectory optimization\cite{sundaralingam2019relaxed}, kinodynamic planning\cite{rus1999hand}, PDDM\cite{nagabandi2020deep}, in-hand object reorientation\cite{chen2022system}, DIME\cite{arunachalam2023dexterous}, DexDeform\cite{li2023dexdeform}}\\
\cmidrule(r){3-3}
{} & {} & \makecell[c]{\textit{Tool Manipulation}: KETO\cite{qin2020keto}, TOG-Net\cite{fang2020learning}, DiffSkill\cite{lin2022diffskill}, tool cognition\cite{tee2022framework}, ATLA\cite{ren2023leveraging}, RoboTool\cite{xu2023creative}}\\
\bottomrule
\end{tabular}
}
\label{tab:embodiedtaskorientedlearning}
\end{table*}

\subsection{Object Grasping}
\label{sec:objectgrasping}
Object grasping is the fundamental cornerstone of object manipulation. It encapsulates a robot's ability to capture targets reliably using end-effectors such as grippers or suction cups. This process requires analyzing object attributes like location, shape, size, and material to formulate grasp strategies that ensure steadfast control while preserving the object's intactness. Grasping methods are further differentiated into single-object grasping \cite{jang2017end} and multi-object grasping \cite{li2024grasp}, each presenting its own set of complexities. Fig. \ref{fig:objectgrasping} illustrates examples of these two types of methods. 

\subsubsection{Single-Object Grasping (SOG)}
Prior research has defined SOG as the configuration of an end-effector designed to achieve partial or complete form-closure or force-closure of a targeted object \cite{lenz2015deep}. Achieving stability and robustness in single-object grasping involves accurately determining object positions and identifying the appropriate grasping pose, which has a wide range of applications in fields such as industrial manufacturing\cite{sun2023novel} and medical assistant\cite{ge2023pixel}.

The typical and direct SOG involves three steps: grasp detection, trajectory planning, and execution. In this pipeline, the robot first captures the local scene using external cameras and plans a set of candidate configurations for the target object. Some methods executed the optimal grasp in an open-loop manner, where the grasp is performed directly without further sensor feedback after selecting the optimal grasp. In open-loop grasping, grasp detection is critical as subsequent steps rely on the coordinates generated during this phase. Consequently, various studies have endeavored to enhance the precision of grasp detection to facilitate effective grasping procedures. For example, Asif \textit{et al.} \cite{asif2017rgb} proposed hierarchical cascaded forests to infer object class and grasp-pose probabilities at both patch and object levels. 
More recently, Zhai \textit{et al.} \cite{zhai2023fanet} designed FANet, which leverages grasp keypoints to enhance the grasp detection accuracy while maintaining real-time efficiency. In AnyGrasp \cite{fang2023anygrasp}, the center-of-mass of objects is incorporated into target detection, and an open-loop strategy is employed throughout the grasping process.

Although open-loop grasping has been extensively studied, it might fail due to inadequate pose estimation and other perception artifacts. To address these issues, closed-loop grasping has been proposed, leveraging real-time feedback to correct perception errors and handle object disturbances. Specifically, object tracking and visual servoing are two primary methods for achieving closed-loop grasping. For instance, Marturi \textit{et al.} \cite{marturi2019dynamic} explicitly tracked the 6DoF object pose and combined it with precomputed grasp poses to enable adaptive grasp planning and execution. Furthermore, Morrison \textit{et al.} \cite{morrison2018closing} proposed GG-CNN to perform close-loop object-independent grasping, using a lightweight CNN to predict pixel-wise grasp quality. After that, Piacenza \textit{et al.} \cite{heppert2024ditto} presented VFAS-Grasp, which uses visual feedback from point clouds and an uncertainty-aware adaptive sampling strategy to maintain a closed-loop system.

\begin{figure}[htbp]
	\centering
	\centerline{\includegraphics[width=1.0\linewidth]{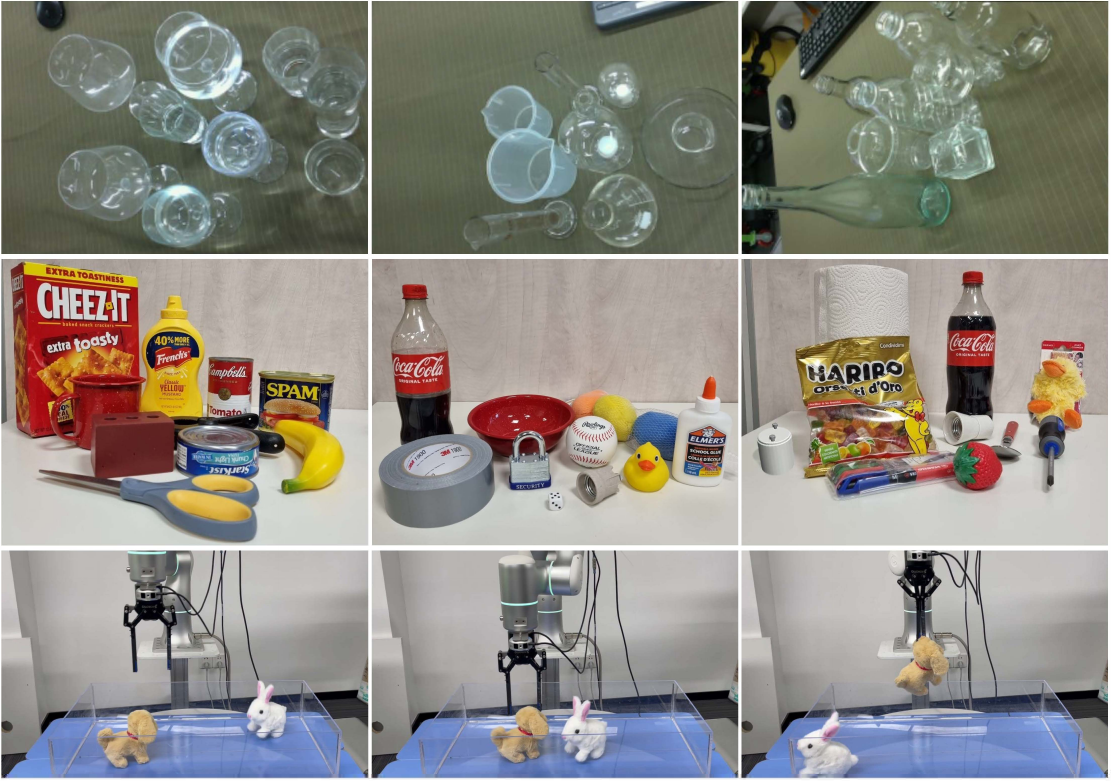}}
	\caption{Illustration of transparent objects (first row), cluttered environment (second row), and dynamic object grasping (last row). The examples are respectively from TRansPose dataset \cite{kim2023transpose}, CEPB benchmark \cite{d2023cluttered}, and Moving GraspNet \cite{liu2023target}.}
	\label{fig:singleobjectgrasping}
\end{figure}

In addition to the general SOG methods mentioned above, three specific tasks, as illustrated in Fig. \ref{fig:singleobjectgrasping}, have garnered significant attention due to their high level of challenge: transparent object grasping, grasping in clutter, and dynamic object grasping.

\textbf{Transparent Object Grasping.}
Transparent objects are items through which light can pass without significant scattering or reflection, such as glass containers and plastic bottles commonly found in daily life. Research into embodied learning technology for grasping transparent objects has a profound impact on robotic applications \cite{jiang2023robotic}. However, grasping transparent objects presents significant challenges. Firstly, the lack of distinctive texture and appearance features, combined with light reflection and refraction, prevents most sensors from accurately capturing surface information, making it difficult for traditional vision systems to recognize and locate these objects. Secondly, the low friction of transparent objects complicates stable manipulation during the grasping process.

Most grasping methods heavily rely on depth images, necessitating precise depth information for transparent objects. Fang \textit{et al.} \cite{fang2022transcg} developed DFNet, an end-to-end depth completion network using RGB images and inaccurate depth maps to produce refined depth maps. Some other approaches utilized NeRFs to generate the depth information of transparent objects directly. For instance, Ichnowski \textit{et al.} \cite{ichnowski2022dex} augmented the specular reflection of transparent objects by placing additional lighting and used NeRFs for transparency-aware depth rendering. However, this method requires several hours of computation per grasp. To speed up the grasping process, Dai \textit{et al.} \cite{dai2023graspnerf} introduced GraspNeRF, which uses six sparse multi-view RGB images for zero-shot NeRF construction and grasp detection, achieving material-agnostic grasp detection in 90ms. In contrast, Lee \textit{et al.} \cite{lee2023nfl} proposed NFL to train neural volume from per-pixel surface normal estimation instead of RGB images and employed segmentation to identify transparent objects, requiring only 40 seconds of training time.

In addition to depth information, some researchers have employed other information to detect transparent objects in the scene. For example, Kim \textit{et al.}\cite{kim2023transpose} created TRansPose, the first large-scale multispectral dataset combining stereo RGB-D, thermal infrared images, and object poses, to promote the study of transparent object grasping. Moreover, Yu \textit{et al.}\cite{yu2023tgf} proposed using TGF-Net to learn surface fragments, edge features, and geometric features for 6D pose estimation of transparent objects, aiming to enhance robustness to variations in appearance. Additionally, they introduced a low-cost dataset generation scheme for obtaining a high-fidelity, large-scale dataset of transparent objects.

\textbf{Grasping in Clutter.}
It refers to a robot's ability to grasp target objects in crowded and cluttered environments precisely, which is crucial for applications like automated home services, manufacturing parts picking, and waste sorting. For instance, in domestic settings, robots must pick up items from a cluttered desk or cabinet for organization or delivery.
Compared to tasks in an organized environment, grasping in clutter is more complex and challenging. This is because target objects may be hidden or overlapping, making them hard to identify and locate, and robots must also avoid collisions with surrounding objects to ensure safety \cite{morrison2020learning}.

Current research primarily focuses on collision-free object grasping \cite{sundermeyer2021contact,wen2022catgrasp}, with the goal of planning a safe and efficient grasping and execution path for the robot to ensure a smooth and unobstructed process. Murali \textit{et al.} \cite{murali20206} proposed a grasp learning method, which uses a deep network called CollisionNet to assess the collision risk of generated grasps in cluttered scenes. Lundell \textit{et al.} \cite{lundell2021ddgc} introduced DDGC, a fast method for generating multi-finger collision-free grasp samples, addressing the issues of long computation times and challenges in obstacle avoidance. Wang \textit{et al.} \cite{wang2021graspness} introduced the concept of graspness, which is a quality metric combining geometric cues and collision labels to evaluate graspable regions in a cluttered scene. To alleviate the reliance on extensive labeled data, Wei \textit{et al.} \cite{wei2023discriminative} introduced a discriminative active learning framework, which employs a discriminator to assess the informational value of unlabeled samples and intelligently select samples for annotation.

Furthermore, a body of research has expanded beyond collision-free object grasping and explored strategies involving grasping and pushing to reposition surrounding objects \cite{xu2021efficient}, which is particularly crucial when the target object is occluded or not directly accessible. Marios \textit{et al.} \cite{kiatos2019robust} addressed the challenge of collision avoidance and proposed using pushing actions to isolate the target object from surrounding clutter. Yang \textit{et al.} \cite{yang2020deep} further explored the problem of grasping invisible objects in clutter and integrated deep Q-Learning with domain knowledge to devise optimal pushing and grasping motions. Recently, Xu \textit{et al.} \cite{xu2023joint} employed a visual-language model to grasp objects in a cluttered environment based on language instructions and utilized a series of obstacle-removal actions to guide the robot to grasp the target object. Although substantial progress has been made, grasping in clutter continues to present significant challenges. Language-conditioned grasping has emerged as a novel and promising research field, increasingly attracting attention for future exploration.

\textbf{Dynamic Object Grasping.}
It is a highly challenging research area focusing on enhancing a robot's ability to grasp moving objects stably. Its applications span from picking up product parts on factory assembly lines to delivering items in household services and even accurately grasping and transferring surgical instruments during surgery procedures. Compared to grasping stationary objects, dynamic object grasping is much more difficult. It requires the robot to quickly adjust its grasp to match the object's movement for precise docking and to predict continuous motion to handle the object's potentially nonlinear and unpredictable trajectories. This presents significant challenges for the robot's real-time processing, adaptability, and advanced motion prediction capabilities.

One key focus of existing work is on human-to-robot (H2R) handover, which aims to enable robots to receive objects from humans. Some studies have been conducted to improve the success rate of H2R handovers by understanding human intention \cite{wang2018controlling,wang2021predicting}. Wang \textit{et al.} \cite{wang2021predicting} explored the integration of multimodal inputs, e.g., vision and language, and proposed a multimodal learning framework to predict human behavioral intentions. An alternative research direction involves dynamic motion planning \cite{yang2021reactive,zhang2023flexible} for H2R handover. Yang \textit{et al.} \cite{yang2021reactive} tackled the challenges posed by object variability in dynamic environments by integrating a closed-loop motion planning strategy with grasp generation. To further enhance the generalizability of H2R handover methods, Wang \textit{et al.} \cite{wang2024genh2r} employed large-scale simulated demonstrations and imitation learning, enabling the robot to pick up objects of any shape transferred by humans in complex trajectories.

Another line of work is human-free moving object grasping, which does not involve human participation. Early studies simplified this problem by assuming prior knowledge of object motion. For example, Ye \textit{et al.} \cite{ye2018velocity} focused on the top-down grasping strategy and proposed a planning algorithm based on analyzing velocity components to predict the trajectory of moving objects. Subsequently, Akinola \textit{et al.} \cite{akinola2021dynamic} expanded on this assumption by developing a dynamic grasping approach that combines reachability and motion awareness, thereby improving the overall success rate without relying on prior knowledge of object motion or constraining the grasping direction. In recent years, the robotics community has expressed significant interest in reactive grasping due to its autonomous adaptability in complex and dynamic environments. Specifically, Liu \textit{et al.} \cite{liu2023target} introduced a target-referenced reactive grasping method that emphasizes both the temporal smoothness and the semantic consistency of the predicted grasp poses. Overall, the technology for grasping moving objects is still evolving, and further in-depth research will enable robots to develop more robust and universally applicable grasping capabilities.

\subsubsection{Multi-Object Grasping (MOG)}
It embodies the advanced performance of robots in efficiently capturing two or more objects within a single operational cycle. This capability holds tremendous potential for various robotic applications, including logistics automation, product packaging, and home services, significantly boosting the efficiency of task execution. Compared to SOG, MOG imposes stricter demands on a robot's comprehensive abilities, encompassing acute perception, sophisticated strategy formulation, and precise execution coordination. Robots must adeptly identify and localize multiple target objects while mastering intricate planning and coordination mechanisms to ensure simultaneous grasping of multiple objects in various environments, achieving an impeccable blend of high efficiency and stability.

Within the domain of MOG, some methods treat it as a holistic grasping problem, in which multiple target objects are regarded as a whole entity for manipulation \cite{agboh2022multi}. These methods intend to encompass all objects with a gripper or fingers in a single action, regardless of their individual placement or stacking configurations. In practice, the contact points are confined to the collective periphery of the objects, and the robot needs to apply precise grasping forces to ensure the ensemble's stability and the grasp's reliability. Sun \textit{et al.} \cite{sun2022multi} developed a comprehensive taxonomy comprising twelve different types of MOG, which incorporate considerations of shape and function. Agboh \textit{et al.} \cite{agboh2023learning} integrated the factor of inter-object friction into their method, significantly improving the robot's ability to grasp multiple objects in a single motion. Many existing methods are grounded in the assumption that the target objects are closely adjacent in space \cite{chen2024multi}. However, this assumption does not always hold in practical, real-world situations. To tackle this challenge, Aeron \textit{et al.} \cite{aeron2023push} introduced the Push-MOG method, which utilizes pushing maneuvers to systematically arrange a disordered ensemble of polygonal objects into compact and easily graspable clusters.

Another type of method involves considering each object as an independent unit, which significantly enhances the robot's flexibility and adaptability. Yao \textit{et al.} \cite{yao2023exploiting} proposed an algorithm that imitates human dexterous grasping, enabling a robotic hand to utilize the cooperative spaces between fingers for efficient and sequential grasping of multiple objects.
Li \textit{et al.} \cite{li2024grasp} introduced the MultiGrasp framework, focusing on maintaining the ability to independently manipulate each object while systematically enhancing the overall grasp efficiency in complex MOG scenarios.
Despite significant progress in this field, there is an urgent need for more exhaustive attention and further in-depth investigation.

\subsubsection{Discussion} SOG, owing to its relative simplicity, has been extensively studied and has achieved significant progress, gradually moving into practical applications. In contrast, MOG, due to its high complexity, has experienced slower progress and still demands further efforts and breakthroughs.

\subsection{Object Manipulation}
\label{sec:objectmanipulation}
\begin{figure}[htbp]
    \centering
    {\includegraphics[width=1.0\linewidth]{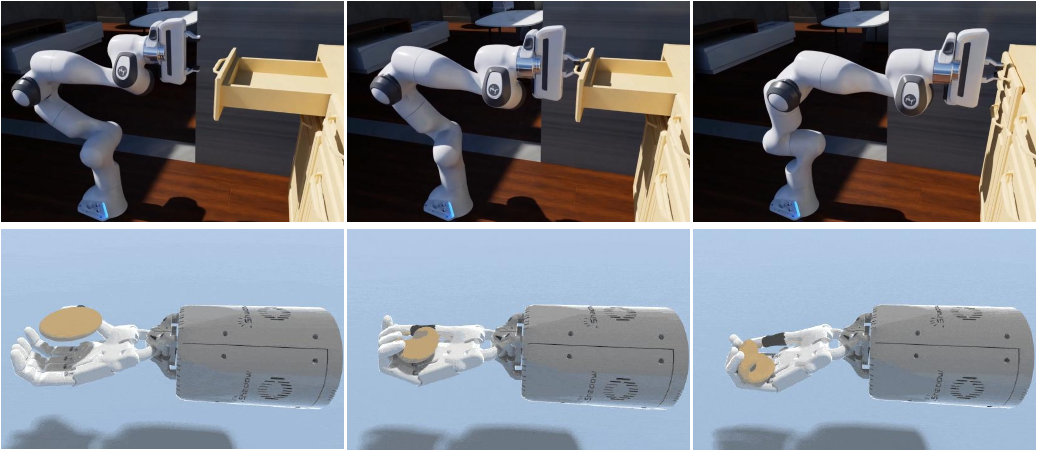}}
    \caption{Illustration of non-dexterous manipulation (top row: close drawer) and dexterous manipulation (bottom row: in-hand manipulation). The examples are respectively from the ARNOLD benchmark \cite{gong2023arnold} and DexDeform \cite{li2023dexdeform}.}
    \label{fig:objectmanipulation}
\end{figure}

Object manipulation involves a wide range of control activities robots perform, from object grasping and utilization to environmental interactions. These capabilities are crucial in various applications, including product assembly, household services, and precision medical surgeries. Currently, the methodologies in this field are divided conceptually into two main categories: non-dexterous manipulation and dexterous manipulation, as depicted in Fig. \ref{fig:objectmanipulation}. Next, we will introduce some representative methods of these two manipulation types.

\subsubsection{Non-Dexterous Manipulation (NDM)}
It refers to using simple end-effectors such as grippers, suction cups, or pushers by robots during task execution instead of relying on delicate finger manipulation or complex hand coordination. This type of manipulation typically has fewer degrees of freedom. It is well-suited for tasks that do not demand high precision or complexity, such as basic gripping, pushing, and pulling. While it may not be as flexible or adaptable as dexterous manipulation, its simplicity and efficiency make it highly promising in fields characterized by repetitive tasks, including industrial assembly, logistics sorting, and agricultural picking.

Pick-and-place is a fundamental task of NDM that has been extensively researched in recent years. It involves a robot picking up objects from one location and placing them in another specified location. Early studies primarily concentrated on estimating the poses of known objects \cite{zhu2014single} in structured environments or relied on scripted planning and motion control \cite{frazzoli2005maneuver}. However, there has been a recent shift towards creating universal pick-and-place policies \cite{berscheid2020self} for novel objects to enhance adaptability across broader scenarios. Furthermore, some research has expanded on the basic pick-and-place capabilities to perform higher-level tasks, such as object rearrangement \cite{zhai2023sg} and kit assembly \cite{zakka2020form2fit}. These advancements represent progress towards more advanced manipulation skills and signify the next steps in complex robotic operations.

Another line of research focuses on improving the intelligence of robots to handle more complex tasks, such as manipulating deformable and articulated objects. For deformable object manipulation, the variability in physical properties across different materials and the complex deformation behaviors under external forces introduce unpredictability in manipulation processes and heighten control complexity. Researchers in this field are drawing insights from human-object interactions in daily life to develop specific manipulation strategies for materials like clothing \cite{huang2022mesh}, ropes \cite{mitrano2021learning}, and fluids \cite{li20223d}. For articulated object manipulation, the core challenge lies in precisely perceiving and controlling each joint part's angle and position while also necessitating a deep understanding of their kinematic properties and dynamic interactions. Typical articulated objects, such as doors, drawers, and buckets, form the nucleus of research interest. The current research frontiers center on establishing benchmarks \cite{geng2023gapartnet,li2024unidoormanip} for articulated object manipulation at the part level and developing universal manipulation policies \cite{geng2023partmanip} that can effectively handle previously unseen shapes and categories.

Although existing methods have demonstrated proficiency in various NDM tasks, they still confront several challenges. For example, in prolonged operations, ensuring high stability and seamless operational continuity is paramount, which requires systems with robust endurance for long-horizon task execution \cite{zhu2022bottom}; in dynamic environments, the robot needs to adapt its position and orientation based on environmental changes or the instantaneous state of objects, which necessitates the integration of active visual adaptation and learning mechanisms \cite{jangir2022look}; when deviations or errors occur during robot operations, the immediate and precise identification of errors coupled with autonomous corrective actions become essential to ensure uninterrupted task completion \cite{liu2024self}. Future research efforts must address these challenges to advance the development of NDM technologies.

\subsubsection{Dexterous Manipulation (DM)}
It aims to replicate subtle human actions, such as unscrewing bottle caps or handling tools. It relies on sophisticated robotic hands \cite{okamura2000overview}, distinct from commonly used parallel grippers in NDM. Typically, these robotic hands emulate the structure of human hands, featuring multiple fingers and exhibiting exceptional flexibility \cite{zhu2023toward}, specializing in precise grasping and manipulation tasks.

Early methods for DM relied on analytical kinematic and dynamic models, using trajectory optimization \cite{sundaralingam2019relaxed} and kinodynamic planning \cite{rus1999hand} to establish robotic control policies and motion trajectories. However, these approaches had a significant limitation as they heavily relied on precise knowledge of dynamic properties and simplified assumptions on object geometries, which are often hard to obtain in complex real-world applications. In recent years, model-based \cite{nagabandi2020deep} and model-free \cite{chen2022system} RL approaches have increasingly become more prevalent in DM. The former aims to train a model from collected data that can predict state transitions and rewards to guide policy optimization. In contrast, the latter does not involve explicit model construction of the environment; instead, it learns directly from experiences gained through interaction with the environment. Another line of work lies in imitation learning, where optimal control strategies are learned from demonstrations \cite{arunachalam2023dexterous}, sometimes integrated with RL to enhance the effectiveness of DM \cite{li2023dexdeform}. These methods have shown effectiveness in executing DM tasks; nonetheless, they are primarily designed and optimized for specific categories of tasks. Consequently, developing universal and broadly adaptable DM frameworks remains an area for further exploration.

\textbf{Tool Manipulation.}
As a universal and fundamental human skill, tool manipulation has emerged as a pivotal focus in the field of DM, which is dedicated to enabling robots to proficiently manipulate a wide range of tools using intricate dexterous hands or specialized end-effectors \cite{qin2023robot}. Its applicability spans from industrial automation to surgical interventions and even space exploration, empowering robots to undertake tasks of remarkable complexity and specificity. In contrast to conventional object manipulation, tool manipulation poses a more stringent challenge to robots. It entails not merely the precise grasping of tools but also the intricate use of tactile feedback to accurately discern the contact status and effects of tool-workpiece interactions \cite{shirai2023tactile}. Considering the wide variety of tools in the real world, with their differing shapes, materials, and usage, robots need to demonstrate robust perception and decision-making capabilities to adapt flexibly and handle the specific physical properties and operational requirements of each tool \cite{zhu2015understanding,saito2021select}.

Current research in tool manipulation often revolves around learning task-specific skills, encompassing movement strategies and manipulation techniques for using tools. This is similar to learning approaches for manipulating non-tool objects but emphasizes integrating tools into robotic actions. Qin \textit{et al.} \cite{qin2020keto} proposed the KETO framework, which employs deep neural networks to predict task-relevant keypoints from point clouds. Fang \textit{et al.} \cite{fang2020learning} introduced TOG-Net (Task-Oriented Grasping Network), which utilizes large-scale simulation for self-supervised learning to optimize tool grasping and manipulation strategies. Lin \textit{et al.} \cite{lin2022diffskill} advanced the field with the DiffSkill framework, leveraging a differentiable physics simulator to learn skill abstraction for tool-based long-horizon manipulation of deformable objects. Distinguished from these methods that rely on prior tool learning, Tee \textit{et al.} \cite{tee2022framework} introduced a framework inspired by neuroscience principles, enabling robots to recognize and deftly apply novel tools to perform a variety of tasks without prior learning. Recently, research trends have ventured into leveraging LLMs to enhance robots' tool manipulation capabilities \cite{ren2023leveraging,xu2023creative}, highlighting a new direction for novel approaches that enable more flexible and efficient robotic tool manipulation.

\subsubsection{Discussion} Both NDM and DM involve diverse and complex tasks. Existing methods are typically designed for several specific tasks and still fall significantly short of achieving truly general object manipulation. 

\begin{table*}[t]
\centering
\caption{Summary of the widely used datasets for robotic manipulation. `Domain' indicates whether the dataset is derived from real-world environments or generated through simulation, with `sim' being short for `simulation'. `-' denotes that the corresponding quantity is unavailable.}
\resizebox{0.8\linewidth}{!}{
\begin{tabular}{clccccc}
\toprule
{\textbf{Task}} & {\textbf{Dataset}} & {\textbf{\#Categories}} & {\textbf{\#Objects}} & {\textbf{Domain}} & {\textbf{Size}} & {\textbf{Modality}}\\
\midrule
\multirow{8}{*}{Object Grasping} & {Cornell \cite{jiang2011efficient}} & - & 240 & real & 885 images, 8,019 grasps & RGB-D\\
{} & {Multi-Object \cite{chu2018real}} & - & $\sim$400 & real & 96 images, 2,904 grasps & RGB-D\\
{} & {Jacquard \cite{depierre2018jacquard}} & - & 11K & sim & 54K images, 1.1M grasps & RGB-D\\
{} & {VR-Grasping-101 \cite{yan2018learning}} & 7 & 101 & sim & 150K grasping demonstrations & RGB-D\\
{} & {ACRONYM \cite{eppner2021acronym}} & 262 & 8,872 & sim & 17.7M parallel-jaw grasps & PointCloud\\
{} & {EGAD \cite{morrison2020egad}} & - & 2,331 & sim & 233K antipodal grasps & PointCloud\\
{} & {GraspNet-1Billion \cite{fang2020graspnet}} & - & 88 & real & 97,280 images, $\sim$1.2B grasps & RGB-D\\
{} & {Grasp-Anything \cite{vuong2023grasp}} & 236 & $\sim$3M & sim & $\sim$1M samples, $\sim$600M grasps & Text/Image\\
\midrule
\multirow{10}{*}{Object Manipulation} & {YCB \cite{calli2017yale}} & 5 & 77 & real & 600 RGB-D images for each object & RGB-D\\
{} & {AKB-48 \cite{liu2022akb}} & 48 & 2,037 & real & 100K generated RGB-D images & RGB-D\\
{} & {PartNet-Mobility \cite{xiang2020sapien}} & 46 & 2,346 & sim & 14,068 articulated parts & PointCloud/RGB-D\\
{} & {GAPartNet \cite{geng2023gapartnet}} & 27 & 1,166 & sim & 8,489 part instances & PointCloud/RGB-D\\
{} & {ManiSkill2 \cite{gu2023maniskill2}} & 20 & 2,000+ & sim & 4M demonstration frames & PointCloud/RGB-D\\
{} & {ARNOLD \cite{gong2023arnold}} & 8 & 1,078 & sim & 10,080 demonstrations & Text/RGB-D\\
{} & {Bi-DexHands \cite{chen2023bi}} & 20 & - & sim & 1,638,400 step demonstrations & Force/PointCloud/RGB-D\\
{} & {DexArt \cite{bao2023dexart}} & 4 & 82 & sim & 6K point clouds for each object & PointCloud\\
{} & {PartManip \cite{geng2023partmanip}} & 6 & 494 & sim & 11 object categories, 1,432 tasks & PointCloud\\
{} & {BEHAVIOR-1K \cite{li2024behavior}} & 1,000 & 9,318 & sim & 50 scenes, 1,949 object categories & RGB-D\\
\bottomrule
\end{tabular}
}
\label{tab:datasets}
\end{table*}

\section{Datasets and Evaluation Metrics}
\label{sec:datasetsandevaluationmetrics}
In this section, we will introduce some primary datasets and evaluation metrics in the area of robotic manipulation.

\subsection{Datasets}
Existing datasets can be divided into two categories based on the differences in specific manipulation tasks: object grasping and object manipulation. Table \ref{tab:datasets} presents an overview of widely used datasets. Most of them are from simulated environments and exhibit considerable variation in categories, objects, data domains, sizes, and modalities. For a detailed description of each dataset, please see Appendix A and B.

\subsection{Evaluation Metrics}
Here, we will introduce some typical evaluation metrics for object grasping and manipulation.

\subsubsection{Object Grasping}
Accuracy is a classic metric for evaluating object grasping, which measures the percentage of correct predictions out of all predicted outputs. There are two critical metrics for determining the correctness of a prediction: the `point' metric \cite{saxena2008robotic} and the `rectangle' metric \cite{jiang2011efficient}. In the `point' metric, a prediction is considered correct if the center point of the predicted rectangle falls within a certain threshold distance from the ground truth grasp point. However, this metric does not account for grasp orientation, which may lead to overestimating actual performance. On the other hand, the `rectangle' metric is designed explicitly for rectangles and incorporates orientation error into the evaluation criteria. It first filters out predicted rectangles with an angular deviation from the ground truth $G$ that surpasses 30 degrees. Then, among the remaining set, it calculates the Intersection over Union (IoU) between the predicted rectangle $\hat{G}$ and $G$:
\begin{equation}
\begin{split}
J(\hat{G}, G) = \frac{|\hat{G}\cap G|}{|\hat{G}\cup G|}.
\end{split}
\label{eq:iou}
\end{equation}
Finally, the prediction $\hat{G}$ is considered correct if $J(\hat{G}, G)$ is greater than a certain threshold $\tau$. 

Besides, Grasp Success Rate (GSR) is commonly adopted as an evaluation metric in real-world robotic experiments. Assuming that a robot performs successful grasp $n$ times out of $m$ grasp attempts, the GSR would be $\frac{n}{m}$. Moreover, several specially tailored evaluation metrics have been proposed, such as completion rate \cite{zhai2023monograspnet} and AP \cite{fang2020graspnet}.

\subsubsection{Object Manipulation}
The most commonly used evaluation metric for object manipulation is the Task Success Rate (TSR). A task is considered successful when it satisfies specific conditions. Generally, each task is performed multiple times using different random seeds to reduce the impact of random variations on assessment results, and the mean value and variance are then reported. The following formula formally defines the TSR:
\begin{equation}
\begin{split}
TSR = \frac{N_s}{N},
\end{split}
\label{eq:tsr}
\end{equation}
in which $N_s$ and $N$ are the number of successful and total executions, respectively. Notably, the conditions for success differ across various types of manipulation tasks. Taking the task of opening a cabinet drawer as an example, the condition for success is that the target drawer has been opened to at least 90\% of its maximum opening range and must remain in a static state \cite{gu2023maniskill2}. There are also some extra metrics to evaluate models from various perspectives, such as the simulated time and kinematic object disarrangement \cite{li2024behavior}. 

\section{Applications}
\label{sec:applications}
With the continuous advancement of artificial intelligence, machine learning, and robotics technology, intelligent robots will be applied more extensively and deeply across various fields. Table \ref{tab:applications} shows the applications of embodied learning for object-centric manipulations, i.e., industrial robots, agricultural robots, domestic robots, surgical robots, and other promising applications. For a detailed description of these applications, please refer to Appendix C.

\begin{table*}[t]
\centering
\caption{Applications of embodied learning for object-centric robotic manipulation.}
\resizebox{\linewidth}{!}{
\begin{tabular}{ccc}
\toprule
{\textbf{Area}} & {\textbf{Example}} & {\textbf{Short Description}}\\
\midrule
\multirow{3}{*}[-1.5ex]{Industrial Robots} & {Assembly line operations \cite{kyrarini2019robot}} & \makecell[l]{Performing tasks such as parts installation and circuit board welding}\\
\cmidrule(r){2-3}
{} & {Packaging and sorting operations \cite{onur2020advanced}} & \makecell[l]{Providing fast and accurate packaging and sorting services for industries like retail and food}\\
\cmidrule(r){2-3}
{} & { Maintenance operations \cite{oyekan2020applying}} & \makecell[l]{Performing equipment maintenance tasks in hazardous environments}\\
\midrule
\multirow{3}{*}[-1.5ex]{Agricultural Robots} & {FarmBot\footnote{\url{https://farm.bot}}} & \makecell[l]{Accomplishing intelligent planting through monitoring plant growth, fertilization, and cultivation control}\\
\cmidrule(r){2-3}
{} & {FAR\footnote{\url{https://www.tevel-tech.com}}} & \makecell[l]{Identify fruits and their ripeness, and performing tasks such as pruning}\\
\cmidrule(r){2-3}
{} & {Rowbot\footnote{\url{https://www.rowbot.com}}} & \makecell[l]{Designing for crops like corn, capable of fertilizing, seeding, and weed removal}\\
\midrule
\multirow{3}{*}[-1.5ex]{Domestic Robots} & {Household assistants \cite{lee2021learning}} & \makecell[l]{Helping with tasks like organizing desks, performing cleaning duties, and folding clothes}\\
\cmidrule(r){2-3}
{} & {Smart caregiving \cite{ye2022rcare}} & \makecell[l]{Offering daily care and health monitoring for people in need of special care}\\
\cmidrule(r){2-3}
{} & {Cooking \cite{liu2022robot}} & \makecell[l]{Automatically completing specific cooking tasks, such as frying eggs, stir-fry, and toasting bread}\\
\midrule
\multirow{3}{*}[-1.5ex]{Surgical Robots} & {Cutting and suturing \cite{lu2021super}} & \makecell[l]{Utilizing specialized surgical tools to perform precise tissue cutting and suturing}\\
\cmidrule(r){2-3}
{} & {Automated control \cite{krishnan2017transition}} & \makecell[l]{Performing intelligent adjustment of the speed and direction of surgical instruments}\\
\cmidrule(r){2-3}
{} & {Surgical collaboration \cite{long2023human}} & \makecell[l]{Collaborating in real-time with doctors during surgery, providing real-time process monitoring and data-driven decision support}\\
\midrule
{Other Applications} & {} & \makecell[l]{Space exploration \cite{wang2023robust}, education \cite{zhong2020systematic}, research \cite{james2020rlbench}}\\
\bottomrule
\end{tabular}
}
\label{tab:applications}
\end{table*}

\section{Challenges and Future Directions}
\label{sec:challengesandfuturedirections}
In the past few years, there has been a significant increase in research on embodied learning methods for object-centric robot manipulation tasks, leading to rapid development in this field. However, current technology still faces some highly challenging issues. Further exploration of these issues will be crucial in promoting the widespread application of intelligent robots in various fields. This section will discuss several challenges and potential future research directions.

\subsection{Sim-to-Real Generalization}
Collecting real-world data for robotic manipulation is difficult, making creating a large-scale dataset challenging. To address this issue, current research primarily focuses on training models within simulation environments \cite{peng2018sim} which offer safe, controllable, and cost-effective learning scenarios, and the ability to generate virtually unlimited simulated training data \cite{aljalbout2024role}. However, real-world environments often present unexpected challenges and variations that simulation environments cannot accurately replicate. This difference might significantly decrease the performance of models trained in simulation environments when applied in real-world situations.

Specifically, the gap between the virtual and real worlds arises from various factors, such as the perception gap, controller inaccuracy, and simulation bias \cite{jiang2024transic}. Recent research has focused on reducing this gap by using methods like domain randomization \cite{muratore2019assessing}, physical constraint regularization \cite{ma2024generalizing}, and iterative self-training \cite{chen2022sim}.
Further studies on this issue can help improve the adaptation of robotic manipulation methods to real-world environments and enhance their performance in practical situations.

\subsection{Multimodal Embodied LLMs} 
\label{subsec:llm}
Humans possess rich perceptual abilities like sight, hearing, and touch, which help them gather detailed information about their surroundings. Besides, humans can utilize learned experiences to perform various tasks. This versatility is also the ultimate goal of general-purpose intelligent robots. To achieve this, robots must be equipped with multiple sensors to perceive the environment and collect multimodal data. Additionally, robots must quickly learn and adapt to new environments and tasks to perform efficient actions. However, this is a significant challenge for intelligent robots.

Recent research has focused on enhancing robots' perception, reasoning, and action-generation abilities using multimodal LLMs \cite{driess2023palm,li2023manipllm,xu2023reasoning,huang2024manipvqa}. For example, Xu \textit{et al.} \cite{xu2023reasoning} introduced a method for tuning reasoning that generates accurate numerical outputs for robotic grasping, leveraging the extensive prior knowledge of LLMs. Huang \textit{et al.} \cite{huang2024manipvqa} integrated affordance and physical concepts into LLMs beyond regular image and text modalities, resulting in better performance in robotic manipulation. These works have promoted the development of multimodal embodied LLMs, but overall, the field is still in its early stages and necessitates further extensive and in-depth research.

\subsection{Human-Robot Collaboration}
Intelligent robots can potentially revolutionize industries such as manufacturing, healthcare, and services. To fully realize this potential, human-robot collaboration is crucial \cite{ajoudani2018progress}. By working together, robots can assist humans, enhancing efficiency and reducing human workload and safety risks. Meanwhile, humans can guide and monitor robot operations to improve accuracy. Nevertheless, achieving perfect human-robot collaboration is challenging due to communication and coordination barriers, over-reliance, and safety issues.

The research community has already achieved some progress in addressing the challenges of human-robot collaboration. For instance, Jin \textit{et al.} \cite{jin2023learning} proposed a two-level hierarchical control framework based on deep RL to establish an optimal human-robot cooperation policy. Wang \textit{et al.} \cite{wang2022co} introduced a policy training method called Co-GAIL, which is based on human-human collaboration demonstrations and co-optimization in an interactive learning process. However, these methods are implemented in simulation environments or can only perform a limited number of tasks, making them unsuitable for practical applications. In the future, human-robot collaboration will remain a crucial research area, requiring continuous exploration to enhance the efficiency and safety.

\subsection{Model Compression and Robot Acceleration}
In applications such as embedded systems, mobile devices, and edge computing, robots with embodied intelligent systems usually have minimal computational resources \cite{chin2020towards}. This makes it essential to optimize and compress the deep models to meet the requirements of storage space, real-time, and accuracy. While LLMs-based methods have made significant advancements in embodied AI, they have also led to increased computational resource demands, posing challenges for implementation on devices with limited computing capabilities. Therefore, future research on model compression is expected to facilitate the practical application of intelligent robots.

In real-world applications, long waiting times often result in a poor user experience. Thus, it is expected that robots should be able to complete tasks quickly. However, many current mainstream models have low operating frequencies. For instance, Google's RT-2 model \cite{zitkovich2023rt} has a decision frequency ranging from 1-5 Hz, depending on the parameter scale of the used VLMs, indicating that there is still a substantial gap before it becomes practical. Recently, the humanoid robot Figure 01\footnote{\url{https://www.figure.ai}} can generate action instructions at a frequency of 200 Hz, which benefits from OpenAI's LLMs and an efficient end-to-end network architecture. This achievement brings greater optimism for future research on robot acceleration.

\subsection{Model Interpretability and Application Safety}
Deep learning-based methods are commonly referred to as ``black boxes'' \cite{wang2023measuring} due to the difficulty in understanding their decision-making process intuitively. For intelligent robots based on deep learning, this black-box characteristic can lead to suspicion and mistrust from users. Particularly in environments where robots interact closely with humans, the lack of transparency can also raise concerns about personal safety \cite{li2019formal}. Therefore, research on the interpretability of embodied learning methods is crucial, which can help people understand the model's decision-making process and increase user trust in robots.

In addition to model interpretability, the safety of intelligent robots needs to be guaranteed from other perspectives, including implementing more reliable online learning and control techniques to prevent potential harm caused by the robot's motion \cite{ji2023safety}. It is also essential to employ adversarial training to protect robots against attacks \cite{jia2022physical} and to design robust safety monitoring methods for detecting possible security risks \cite{machin2016smof}. Further research in these areas is expected to improve robots' safety and reliability in practical applications.

\section{Conclusion}
\label{sec:conclusion}
In this paper, we present a comprehensive survey of the existing methods for embodied learning in object-centric robotic manipulation. We begin by introducing the concept of this task and its essential components and then compare it with related survey articles. Next, we systematically present the main works across three categories. We then explore the commonly used datasets and evaluation metrics, highlighting some representative applications. Finally, we discuss the challenges and suggest promising directions for future research. We hope this survey will provide researchers with a comprehensive understanding and new insights in this emerging field.


\bibliographystyle{IEEEtran}
\bibliography{refs}

\newpage
\section*{Appendix}
\subsection{Datasets for Object Grasping}
The objective of the object grasping task is for the robotic arm to be able to successfully and stably grasp objects. Therefore, the related datasets are mainly constructed around aspects such as enriching the categories of objects and the grasping poses.

The \textbf{Cornell} dataset \cite{jiang2011efficient} encompasses 240 different objects, for which 885 RGB-D images have been captured with a top-down imaging perspective. It provides 8,019 manually annotated positive and negative grasp samples. This dataset was the first to introduce oriented rectangles to represent grasps, enabling efficient training of grasp detection network models on images.

The \textbf{Multi-Object} dataset \cite{chu2018real} comprises 96 real-world images, each containing 3-5 objects, amounting to approximately 400 objects in total. Multiple grasps have been annotated for each object, with a total of 2,904 grasps included, making it suitable for evaluating multi-object/multi-grasp tasks.

The \textbf{Jacquard} dataset \cite{depierre2018jacquard} contains 54K images of 11K objects and provides 1.1 million grasping positions, which are represented by 2D rectangles on the images. All data are generated through simulations on CAD models and can be leveraged for image-based grasping position estimation.

The \textbf{VR-Grasping-101} dataset \cite{yan2018learning} includes 101 daily objects across 7 categories and synthesizes approximately 150K grasping samples based on human demonstrations collected in Virtual Reality (VR). This dataset is particularly suitable for 6DoF parallel-jaw grasping tasks.

The \textbf{ACRONYM} dataset \cite{eppner2021acronym} is constructed based on simulation, comprising 8,872 objects across 262 categories from the ShapeNetSem dataset \cite{savva2015semantically}, and provides data on 17.7 million parallel-jaw grasping poses. As a large-scale grasping dataset, it can be used to train learning-based grasp detection algorithms.

The \textbf{EGAD} dataset \cite{morrison2020egad} contains 2,331 objects represented by 3D meshes, with each object annotated with 100 antipodal grasps. Furthermore, a curated selection of 49 objects amenable to 3D printing has been chosen to facilitate the testing of robotic grasping tasks in real-world scenarios.

The \textbf{GraspNet-1Billion} dataset \cite{fang2020graspnet} encompasses 88 daily objects and has collected 97,280 RGB-D images of these objects from 190 cluttered scenes, along with providing accurate 3D mesh models. All data were acquired using real-world sensors and cameras. Furthermore, over one billion grasp poses have been annotated through analytic computation, offering a large-scale benchmark for the advancement of robotic object grasping techniques.

The \textbf{Grasp-Anything} dataset \cite{vuong2023grasp} is constructed based on foundational models and includes 236 object categories from the LVIS dataset \cite{gupta2019lvis}. It provides one million scene descriptions generated by ChatGPT and corresponding images produced by Stable Diffusion \cite{rombach2022high}, with a total object count of approximately 3 million. For each object, there is an average of 200 grasps represented by 2D rectangles, amounting to roughly 6 million grasps in total. This dataset has the potential to substantially bolster research in the domain of zero-shot grasp detection.

\subsection{Datasets for Object Manipulation}
Compared to grasp detection, object manipulation is a more complex and general task, encompassing both the action of grasping and the various operations performed on objects after grasping. Therefore, datasets in this field focus more on covering a wider variety of task types, scenarios, and skills.

The \textbf{YCB} dataset \cite{calli2017yale} consists of 5 types of object sets, totaling 77 sub-classes of objects. All data were captured using 5 RGB-D sensors and 5 high-resolution RGB cameras. For each object, it provides 600 RGB-D images, 600 high-resolution RGB images, segmentation masks, calibration information, and 3D mesh models with texture mapping. This dataset is primarily useful for research related to robotic grasping and manipulation.

The \textbf{AKB-48} dataset \cite{liu2022akb} contains 48 types of objects, encompassing a total of 2,037 articulated object instances. Each instance has been scanned from the real world and manually refined by humans. Based on these models, 100K RGB-D images have been generated for network training. In addition, 10K real-world images have been collected, with 50\% used for fine-tuning and the other 50\% for testing. This dataset can be utilized to facilitate the generalization of robotic manipulation methods from simulation to reality.

The \textbf{PartNet-Mobility} dataset \cite{xiang2020sapien} encompasses 46 categories of common indoor objects, totaling 2,346 articulated object models, and provides 14,068 annotations of moving parts. Specifically, the motion of parts is categorized into three types: hinge, slider, and screw. This dataset serves as an evaluation benchmark for robotic perception and part-based manipulation tasks.

The \textbf{GAPartNet} dataset \cite{geng2023gapartnet} is a large-scale part-centric dataset for object manipulation, featuring 1,166 articulated object models across 27 categories, all sourced from the AKB-48 dataset and PartNet-Mobility dataset. It defines 9 classes of cross-category GAPart and provides annotations for functional parts on each object, amounting to a total of 8,489 GAPart instances. This dataset is suitable for domain-generalizable tasks on object perception and manipulation.

The \textbf{ManiSkill2} dataset \cite{gu2023maniskill2} is developed on the OpenAI Gym simulator \cite{brockman2016openai} and consists of 20 manipulation tasks that span a diverse range of task types, including stationary/mobile, single/dual-arm, and rigid/soft. It contains over 2,000 object models and offers more than 4 million demonstration frames. This dataset is particularly well-suited for advancing and evaluating research in the field of learning generalizable manipulation skills.

The \textbf{ARNOLD} dataset \cite{gong2023arnold} is built on the Isaac Gym simulator \cite{makoviychuk2021isaac} and encompasses 8 language-grounded manipulation tasks, each presenting four goal states articulated through human language. This dataset provides a comprehensive collection of 10,080 learning demonstrations, with each demonstration consisting of 4-6 keyframes. It consists of 1,078 3D object models across 40 distinct categories and 1,114 scenes spanning 20 different types. This dataset establishes an evaluation benchmark for methods of language-conditioned manipulation and the generalization of learned skills.

The \textbf{Bi-DexHands} dataset \cite{chen2023bi} includes 20 types of bimanual manipulation tasks, features thousands of target objects, and provides multi-modal information like contact force, RGB image, RGB-D image, and point cloud. This dataset offers a comprehensive benchmark for general reinforcement learning approaches aimed at dexterous two-handed manipulation tasks.

The \textbf{DexArt} dataset \cite{bao2023dexart} comprises 4 categories of objects, i.e., faucet, bucket, laptop, and toilet, with a total of 82 objects included. For each object, it provides 6k point clouds, which encompass both the actual observed and imagined points. The primary focus of this dataset is on the generalizable dexterous manipulation oriented towards articulated objects.

The \textbf{PartManip} dataset \cite{geng2023partmanip} is constructed based on the GAPartNet dataset and comprises 494 objects across 11 different types. It consists of 1,432 sub-tasks that fall under 6 major task categories, respectively OpenDoor, OpenDrawer, CloseDoor, CloseDrawer, PressButton, and GraspHandle. This dataset is designed to advance the research into part-based cross-category object manipulation methods.

The \textbf{BEHAVIOR-1K} dataset \cite{li2024behavior} encompasses a comprehensive collection of 1,000 behavioral categories derived from 50 everyday scenarios, incorporating 1,949 object classes with a total of 9,318 individual object models. Additionally, it provides rich physical and semantic annotations for each object. All data within BEHAVIOR-1K are generated within a simulation environment named OMNIGIBSON, which is particularly tailored for the evaluation of diverse approaches to complex robotic manipulation tasks.

\subsection{Description of Applications}
\subsubsection{Industrial Robots}
Traditional industries, represented by manufacturing, rely heavily on human resources. However, intelligent robots are expected to revolutionize the conventional industrial production model, achieving goals such as increasing production efficiency, reducing labor costs, and enhancing safety. Typical application scenarios include 1) Assembly line operations \cite{kyrarini2019robot}, where robots can perform tasks such as parts installation and circuit board welding; 2) Packaging and sorting operations \cite{onur2020advanced}, where robots can provide fast and accurate packaging and sorting services for industries like retail and food; 3) Maintenance operations \cite{oyekan2020applying}, where robots can perform equipment maintenance tasks in hazardous environments.

In the process of robots becoming increasingly intelligent but not yet fully autonomous, collaboration between robots and humans 
is essential in many complex and high-precision task scenarios. Additionally, robots may encounter operational anomalies or make mistakes, which requires the implementation of intelligent detection and fault diagnosis methods \cite{chen2023compound}. This is crucial for ensuring intelligent robots' stable and reliable application within the industrial field.

\subsubsection{Agricultural Robots}
In modern agriculture, intelligent robots are crucial in completing various tasks within farms and orchards, such as planting, nurturing, and harvesting crops. This promotes high-quality and sustainable agricultural development \cite{shvets2024robotics}. Some representative agricultural intelligent robots include 1) FarmBot\footnote{\url{https://farm.bot}}, which accomplishes intelligent planting through monitoring plant growth, fertilization, and cultivation control; 2) FAR\footnote{\url{https://www.tevel-tech.com}}, which utilizes artificial intelligence and computer vision technology to identify fruits and their ripeness, and can perform tasks such as pruning; 3) Rowbot\footnote{\url{https://www.rowbot.com}}, designed for crops like corn, capable of fertilizing, seeding, and weed removal.

In agricultural settings, goods are often delicate and prone to damage \cite{van2021current}, as seen in tomato harvesting. Applying too much force can harm the tomatoes, while too little force can cause them to slip and fall. This makes it challenging to achieve precise control and gentle handling. Additionally, robots may encounter obstacles such as branches and leaves in the open agricultural environment,
requiring high positional accuracy and flexibility during operation.

\subsubsection{Domestic Robots}
Domestic intelligent robots have promising applications in areas such as family assistance, caregiving, and household chores. They are capable of enhancing living quality and convenience and providing assistance to individuals with particular needs. Specifically, some valuable application scenarios include 1) Household assistants \cite{lee2021learning}, which help with tasks like organizing desks, performing cleaning duties, and folding clothes, thereby alleviating the daily household workload; 2) Smart caregiving \cite{ye2022rcare}, which offers daily care and health monitoring for people in need of special care; 3) Cooking \cite{liu2022robot}, capable of automatically completing specific cooking tasks, such as frying eggs, stir-fry, and toasting bread.

Due to the vast diversity in home environments and the high complexity of tasks, it is essential for robots to quickly adapt to different household settings and handle potentially unexpected situations as well as new types of tasks. Taking cooking as an example, it involves a wide variety of ingredients and vastly different cooking steps
, and the process also requires continuous identification of the food's state \cite{kawaharazuka2024continuous}, which poses a significant challenge for robots. Another critical aspect is that such robots must be sufficiently safe and reliable, especially around children. Furthermore, cost is one of the significant factors affecting the widespread adoption of domestic robots. Consequently, identifying strategies to reduce the expense of robots while ensuring the quality of service is also a crucial issue that must be considered.

\subsubsection{Surgical Robots}
Research on robots in the surgical field is rapidly advancing. These robots have the potential to serve as intelligent assistant tools to help doctors improve the quality and efficiency of surgery. Some typical application scenarios include 1) Cutting and suturing \cite{lu2021super}, where robots utilize specialized surgical tools to perform precise tissue cutting and suturing, thereby reducing surgical trauma and medical staff workload; 2) Automated control \cite{krishnan2017transition}, where intelligent adjustment of the speed and direction of surgical instruments is performed, enabling precise control over surgical progress and outcomes; 3) Surgical collaboration \cite{long2023human}, where robots collaborate in real-time with doctors during surgery, providing real-time process monitoring and data-driven decision support.

The privacy of surgical data makes obtaining large-scale real-world data challenging.
Current methods often rely on simulation environments to enhance surgical machine learning \cite{yu2024orbit}. However, there is a significant gap between simulation and reality, challenging intelligent robots to make real progress in practical applications. Therefore, while intelligent robots can perform certain parts of surgical procedures, it will take considerable time before they can replace doctors. In particular, complex and enduring surgical tasks still require doctors' oversight to ensure the surgery's safety and effectiveness.

\subsubsection{Other Applications}
In addition to the applications above, intelligent robots can also be used in fields such as space exploration \cite{wang2023robust}, education \cite{zhong2020systematic}, and research \cite{james2020rlbench}. The skills required in these fields are different, and most of the existing work involves designing specialized intelligent models tailored to domain expert knowledge. While these methods perform well on specific tasks, their generalization capability could be improved. Fortunately, recent explorations based on general large foundation models offer a promising solution.

 



\begin{IEEEbiographynophoto}{Ying Zheng} is currently a Research Fellow at The Hong Kong Polytechnic University, Hong Kong, China. Before that, he received the Ph.D. degree from Harbin Institute of Technology, Harbin, China. His research interests include computer vision and embodied AI.
\end{IEEEbiographynophoto}

\vspace{-22pt}

\begin{IEEEbiographynophoto}{Lei Yao}
received the B.Eng. and M.Eng. degrees from Huazhong University of Science and Technology, Wuhan, China, in 2020 and 2023, respectively. He is currently working towards the Ph.D. degree with The Hong Kong Polytechnic University. His research interests include 3D scene understanding and Embodied AI.
\end{IEEEbiographynophoto}

\vspace{-22pt}

\begin{IEEEbiographynophoto}{Yuejiao Su}
received the B.Sc. and M.Sc. degrees from the Northwestern Polytechnical University in 2020 and 2023 respectively. She is currently working towards the Ph.D. degree with The Hong Kong Polytechnic University. Her research interests include Egocentric analysis, image segmentation, and embodied AI.
\end{IEEEbiographynophoto}

\vspace{-22pt}

\begin{IEEEbiographynophoto}{Yi Zhang}
received the B.Sc. and M.Sc. degrees from the Hong Kong University of Science and Technology and Institut Polytechnique de Paris in 2020 and 2022 respectively. She is currently working towards the Ph.D. degree with The Hong Kong Polytechnic University. Her research interests include 3D scene understanding and embodied AI.
\end{IEEEbiographynophoto}

\vspace{-22pt}

\begin{IEEEbiographynophoto}{Yi Wang}
(Member, IEEE) received B.Eng. degree in electronic information engineering and M.Eng. degree in information and signal processing from the School of Electronics and Information, Northwestern Polytechnical University, Xi’an, China, in 2013 and 2016, respectively. He earned PhD in the School of Electrical and Electronic Engineering from Nanyang Technological University, Singapore, in 2021. He is currently a Research Assistant Professor at the Department of Electrical and Electronic Engineering, The Hong Kong Polytechnic University, Hong Kong. His research interest includes Image/Video Processing, Computer Vision, Intelligent Transport Systems, and Digital Forensics.
\end{IEEEbiographynophoto}

\vspace{-22pt}

\begin{IEEEbiographynophoto}{Sicheng Zhao}
(Senior Member, IEEE) received the Ph.D. degree from the Harbin Institute of Technology, Harbin, China, in 2016. He was a Visiting Scholar with the National University of Singapore, Singapore, from 2013 to 2014, a Research Fellow with Tsinghua University, Beijing, China, from 2016 to 2017, a Postdoc Research Fellow with the University of California at Berkeley, Berkeley, CA, USA, from 2017 to 2020, and a Postdoc Research Scientist with Columbia University, New York, NY, USA, from 2020 to 2022. He is currently a Research Associate Professor with Tsinghua University. His research interests include affective computing, multimedia, and computer vision. He is an associate editor of IEEE TIP and IEEE TAFFC.
\end{IEEEbiographynophoto}

\vspace{-22pt}

\begin{IEEEbiographynophoto}{Yiyi Zhang}
received her B.Eng. degree from Zhejiang University and M.Sc. degree from the Institut Polytechnique de Paris - Télécom Paris. She is currently working towards the Ph.D. degree with The Chinese University of Hong Kong. Her research interests include Computer Vision and Medical AI.
\end{IEEEbiographynophoto}

\vspace{-22pt}

\begin{IEEEbiographynophoto}{Lap-Pui Chau}
 (Fellow, IEEE) received a Ph.D. degree from The Hong Kong Polytechnic University in 1997. He was with the School of Electrical and Electronic Engineering, Nanyang Technological University from 1997 to 2022. He is currently a Professor in the Department of Electrical and Electronic Engineering, The Hong Kong Polytechnic University. His current research interests include image and video analytics and autonomous driving. He was the chair of Technical Committee on Circuits \& Systems for Communications of IEEE Circuits and Systems Society from 2010 to 2012. He was general chairs and program chairs for some international conferences. Besides, he served as associate editors for several IEEE journals and Distinguished Lecturer for IEEE BTS.
\end{IEEEbiographynophoto}

\vfill

\end{document}